\documentclass[journal,10pt]{IEEEtran}
\usepackage{blkarray}                                      % to support matrix
\usepackage{algpseudocode}                                 % new algorithm package
\usepackage{algorithm}
\usepackage{graphicx}                                      % include pdf figures
\usepackage{amsmath}
\usepackage{amssymb}
\usepackage{amsfonts}
\usepackage{amsthm}
\usepackage[mathcal]{eucal}
\usepackage{booktabs}
\usepackage{enumerate}
\usepackage{multirow}
\usepackage{kbordermatrix}
\usepackage[subrefformat=parens,farskip=0pt,justification=centering]{subfig}
\usepackage{color}
\usepackage{cite}                                          % more citations in one bracket
\usepackage{comment}                                       % use comment
\usepackage{soul}                                          % use highlight command \hl{}
\soulregister\cite7
\soulregister\ref7
\soulregister\pageref7
\usepackage{etoolbox}                                      % commands \newtoggle, \toggletrue, \iftoggle
\usepackage{url}
\usepackage{nth}                                           % nth command
\usepackage{bm}                                            % bm command
\usepackage{courier}
\usepackage{balance}
\usepackage{threeparttable}
\usepackage[bookmarks=false]{hyperref}
\hypersetup{
    colorlinks = true,
    citecolor  = blue,
    linkcolor  = blue,
    urlcolor   = blue,
}
\usepackage{tikz}
\usetikzlibrary{positioning,calc,fit,decorations.pathmorphing}
\usepackage{filecontents}                                  % support to pgfplots
\usepackage{pgfplots}
\usepackage{pgfplotstable}
\usepackage{scalefnt}
\pgfplotsset{compat=newest}
\usepackage{caption}
\usepackage{cleveref}
\Crefformat{figure}{Fig.~#2#1#3}                           % "Fig.", instead of "Figure"
\Crefname{subfigure}{Fig.}{Figs.}
\Crefname{figure}{Fig.}{Figs.}
\Crefformat{table}{TABLE~#2#1#3}                           % "TABLE", instead of "Table"
\definecolor{CUHKorange}{RGB}{244,106,18} %F47012
\definecolor{CUHKblue}{RGB}{0,111,190}    %006FBE
\definecolor{CUHKgreen}{RGB}{0,127,128}   %007F80
\definecolor{CUHKred}{RGB}{228,46,36}     %E42E24
\definecolor{CUHKyellow}{RGB}{198,148,34} %C69422
\definecolor{CUHKdark}{RGB}{114,44,114}   %722C72
\definecolor{CUHKmiddle}{RGB}{144,44,144} %902C90
\definecolor{CUHKlight}{RGB}{167,44,167}

\renewcommand{\vec}[1]{\boldsymbol{#1}}    % re-define vec command

\usepackage{tcolorbox}
\tcbuselibrary{skins,breakable}
    {\endtcolorbox}

\newtheorem{myproblem}{\textbf{Problem}}
\newtheorem{mydefinition}{\textbf{Definition}}

\algrenewcommand\textproc{\texttt}

\makeatletter
\let\OldStatex\Statex
\renewcommand{\Statex}[1][3]{%
  \setlength\@tempdima{\algorithmicindent}%
  \OldStatex\hskip\dimexpr#1\@tempdima\relax
}
\makeatother

\RequirePackage[normalem]{ulem} %DIF PREAMBLE
\RequirePackage{color}\definecolor{RED}{rgb}{1,0,0}\definecolor{BLUE}{rgb}{0,0,1} %DIF PREAMBLE
    {\endtcolorbox}
\graphicspath{{./figs/}}

\usepgfplotslibrary{groupplots}
\usetikzlibrary{patterns}

\definecolor{myorange}{RGB}{244,106,18} %F47012
\definecolor{myblue}{RGB}{0,111,190}    %006FBE
\definecolor{mygreen}{RGB}{0,127,128}   %007F80
\definecolor{myred}{RGB}{228,46,36}     %E42E24
\definecolor{myyellow}{RGB}{198,148,34} %C69422
\definecolor{mydark}{RGB}{114,44,114}   %722C72
\definecolor{mymiddle}{RGB}{144,44,144} %902C90
\definecolor{mylight}{RGB}{167,44,167}  %A72CA7

\begin{document}

\twocolumn
\setcounter{page}{1}
	
	\title{
		Graph Attention-Based Symmetry Constraint Extraction for Analog Circuits
	}
	
	\author{
    Qi Xu,
    Lijie Wang,
    Jing Wang,
    Lin Cheng,
    Song Chen,
    Yi Kang
    \thanks{This work was supported by the National Natural Science Foundation of China (NSFC) under grant No. 62141415 and USTC Research Funds of the Double First-Class Initiative under grant No. YD2100002012. (Corresponding authors: Song Chen, Jing Wang.)}
    \thanks{Q.~Xu, L.~Wang, J.~Wang, L.~Cheng, S.~Chen and Y.~Kang are with School of Microelectronics, University of Science and Technology of China, Hefei, China. (e-mail: \href{mailto:xuqi@ustc.edu.cn}{\texttt{xuqi@ustc.edu.cn}}, \href{mailto:songch@ustc.edu.cn}{\texttt{songch@ustc.edu.cn}}, \href{mailto:jw2019@ustc.edu.cn}{\texttt{jw2019@ustc.edu.cn}})}
	}
	%\markboth{IEEE TRANSACTIONS ON COMPUTER-AIDED DESIGN OF INTEGRATED CIRCUITS AND SYSTEMS, VOL. , NO.}{Lijie Wang et al.}

    \maketitle

    \begin{abstract}
In recent years, analog circuits have received extensive attention and are widely used in many emerging applications.
The high demand for analog circuits necessitates shorter circuit design cycles. %has necessitated the need to shorten the circuit design cycle.
To achieve the desired performance and specifications,
various geometrical symmetry constraints must be carefully considered during the analog layout process.
%In particular, the analog circuit design is critical to achieving high-quality layouts, which often require constraints such as symmetry and matching.
However, the manual labeling of these constraints by experienced analog engineers is a laborious and time-consuming process.
%Thus, it is necessary to automatically extract the constraints of analog circuit to reduce the design procedure. %there is a need for automatic extraction of analog circuit constraints to shorten the design cycle.
To handle the costly runtime issue,
we propose a graph-based learning framework to automatically extract symmetric constraints in analog circuit layout.
The proposed framework leverages the connection characteristics of circuits and the devices' information to learn the general rules of symmetric constraints,
which effectively facilitates the extraction of device-level constraints on circuit netlists.
The experimental results demonstrate that compared to state-of-the-art symmetric constraint detection approaches,
our framework achieves higher accuracy and {F$_1$-score}.
%leading to shorter analog circuit design cycles and better circuit performance.
\end{abstract}

	\section{Introduction}
\label{sec:intro}
\IEEEPARstart{T}{he} demands for analog integrated circuits (ICs) are increasing rapidly in various fields,
such as consumer electronics, medical electronics, smart cars, and other emerging applications.
The growing demand for analog ICs necessitates an expedited design process.
However, despite some progress in electronic design automation (EDA) tools for analog ICs~\cite{xu2019magical,kunal2019align},
they still encounter challenges in meeting design requirements during the time-consuming and laborious layout process.
To ensure optimal functionality and performance, well-defined constraints (symmetry, matching, etc.) need to be satisfied to guide the layout design.
%To ensure optimal functionality and performance,
%well-defined constraints (symmetry, matching, etc.) need to be satisfied to guide the layout design process.
%In analog circuit design automation,
%advanced analog layout synthesis tools rely heavily on well-defined constraints
%ensuring optimal functionality and performance.
%These constraints include symmetry, network shielding and minimal parasitic,
%The symmetric constraint is one of the most representative constraints in analog circuits.
For example, differential topology is often used in analog circuit design to reject common-mode noise and enhance the robustness of the circuit.
%In order to prevent these topological devices from mismatching due to size or asymmetry in the layout design and degrade the performance of the circuit,
To prevent the performance degradation of the layout due to the asymmetry of these topological devices,
it is necessary to annotate the constraints in advance.
%\Cref{fig0} shows a layout example of a bandgap reference circuit with and without satisfying symmetric constraints.
%\Cref{fig:00} and \Cref{fig:01} show the different results of the layout of the same Bandgap circuit with or without symmetric matching.
%In terms of temperature coefficient (TC) and power supply rejection ratio (PSRR),
%power suppression ratio (PSRR),
%larger absolute values represent better performance,
%indicating that symmetric matching results are more advantageous.
%As the number of asymmetric pairs increases, it undoubtedly further degrades the overall performance.
%Thus, it is necessary to conduct layout constraint extraction to guide the layout design for functionality and performance.

Over the past decade, progress on automated analog IC layout design tools has been relatively slow,
with many tools relying on the domain knowledge of experienced engineers.
As a result, analog IC layout design still remains a highly manual and expensive task,
especially in the face of complex and diverse analog circuits with high design flexibility.
Researchers have proposed sensitivity analysis-based methods to detect constraints~\cite{choudhury1991constraint,malavasi1996automation,charbon1993generalized}.
Using circuit simulation and sensitivity analysis, the key devices that impact performance are identified,
and then constraint conditions are generated based on these devices.
Although the sensitivity analysis-based approaches can automatically identify critical constraints,
due to the computational effort and time-consuming simulation of sensitivity analysis,
%these methods are limited to small system-level circuits.
these methods are limited to standard analog circuit modules with a limited number of devices, such as a simple OTA circuit.
%and are difficult to be applied to larger circuits,
%especially for non-linear circuits that require significant changes in sensitivity.
Moreover, some works~\cite{2004Hierarchical,Lin2015A,2015Analog} attempt to convert circuit netlists into graphs and search for subgraphs in a designed database to infer the constraint matching for new designs.
However, they require a sufficient number of valid and accurate circuits in the database, and thus are not suitable for increasingly complex analog circuits.
In addition, by adopting signal flow analysis to convert analog circuits into simple bipartite graphs,
symmetric constraints can be extracted through graph isomorphism algorithms,
and matching constraints are further identified through primitive cell recognition with signal flow analysis~\cite{hao2004constraints,zhou2005analog,massier2008sizing,eick2011comprehensive}.
Similarly, an S$^3$DET flow is presented to detect system symmetry constraints by leveraging spectral graph analysis and graph centrality~\cite{liu2020s}.
However, these methods strongly rely on the similarity threshold parameters and face generalization challenges.
%, adjacent circuit topologies of subcircuits based on the centrality of the graph, and measures the similarity of the graph through spectral analysis to find a match. However, these methods strongly rely on the selection of similarity threshold parameters, and may also face challenges in promotion and generalization.

%\begin{figure}[tb!]
%	\centering
%	\subfloat[]{\includegraphics[height=0.85in]{figs/matched.png} \label{fig:00}}
%   \hspace{.1in}
%	\subfloat[]{\includegraphics[height=0.85in]{figs/unmatcehd.png} \label{fig:01}}
%	\caption{Layout of a bandgap reference circuit with (a) accurate symmetric constraints and (b) a symmetric constraint removed.} %Matched Bandgap and UnMatched Bandgap.}
%	\label{fig0}
%\end{figure}

%Over the past decade,
Recently, Artificial Intelligence (AI) technology has been widely used in analog ICs.
%With the development of artificial intelligence (AI) technology,
%there have been notable developments in the realm of analog ICs.
For example, AI has enhanced the conventional approach~\cite{guo2023open} for fault detection in analog circuits~\cite{gao2023incipient,gao2022novel},
thereby improving the reliability of analog ICs.
Similarly, AI-related techniques are presented for analog layout placement~\cite{dhar2021fast,zhu2022tag} and the performance prediction of analog circuits~\cite{hakhamaneshi2022pretraining}.
%while the paper~\cite{hakhamaneshi2022pretraining} focuses on the performance prediction of analog circuits.
Recently, leveraging AI methods toward automated extraction of layout constraints for analog ICs is emerging as a trend.
%there is a trend toward automated extraction of layout constraints for analog ICs based on deep learning algorithms.
Kunal \textit{et al}.~\cite{kunal2020general} utilize a graph neural network (GNN) to handle multiple levels of symmetry hierarchies.
%\textcolor{blue}{The graph neural network (GNN) is employed to handle multiple levels of symmetry hierarchies~\cite{kunal2020general}.}
%It first represents the net and device instances as nodes of a circuit graph,
%and then candidate node pairs are detected by traversing the graph.
%If the parameters of the candidate pairs are the same,
%the framework recognizes them as symmetry.
However, the function of the GNN in the work is only to solve the graph edit distance (GED) to measure graph similarity,
ignoring the adjacent topology at the device-level,
and thus cannot effectively extract device-level constraints.
%However, the focus of this work is on using neural networks to solve the graph editing distance (GED) for judging graph similarity,
%ignoring the adjacent topology at the device level, and thus unable to extract device-level constraints.
%The work~\cite{gao2021layout} presents a simple and effective methodology to identify symmetric matching.
A simple and effective methodology is developed in~\cite{gao2021layout} for identifying symmetric matching.
It first represents device instances and corresponding pins in the circuit as nodes of a circuit graph,
and then embeds the types of devices and pins as node features.
Through training a GraphSAGE model~\cite{hamilton2017inductive},
adjacent information is aggregated into node embeddings, which are adopted to identify whether two nodes are symmetric.
Although the approach improves the accuracy of symmetry constraint extraction,
it cannot detect the unpaired constraints with more than two devices,
which is not reasonable in analog circuits.
%Besides, Chen \textit{et al}.~\cite{chen2021universal} propose a gated graph neural network (GGNN)-based method leveraging unsupervised learning to detect symmetric constraints in analog circuits.
Chen \textit{et al}.~\cite{chen2021universal} further propose a gated graph neural network (GGNN)-based method leveraging unsupervised learning to detect symmetric constraints in analog ICs.
Since the GGNN aggregates neighboring node features with the convolution operation,
the final embedding of each node contains information from its neighboring sub-graph.
Based on the node embedding, the cosine similarity of the two nodes is computed as a criterion for symmetry.
%But the circuit netlist is converted as a heterogeneous directed multigraph,
%and the computation overhead of the GGNN is tremendously high.
However, the circuit netlist is converted as a heterogeneous directed multigraph, and thus the computation overhead of the GGNN is tremendously high.
Furthermore, the adopted edge features only consider the connections of MOSFETs and passive devices,
making it challenging to apply to other types of devices.
Additionally, a comprehensive summary of recent advancements in constraint extraction is provided in~\cite{zhu2022automating}.
Besides, integrating GNN with traditional graph-based algorithms achieves the extractions of analog layout constraints at different design levels~\cite{kunal2023gnn}.
%the use of GNN and graph-based algorithms~\cite{kunal2023gnn} to extract analog layout constraints at different design levels has also yielded promising results.
Differing from the above approaches of extracting information directly from SPICE netlists,
a novel constraint extraction method based on the layout templates is recently proposed in~\cite{yao2023automatic}.
However, the approach requires a large number of previous high-quality layouts.
%\textcolor{blue}{However, the edge features of the network fail to consider devices beyond MOS transistors and passive devices. Furthermore, the handling of node features lacks universality, casting doubt on the method's generalizability.}
%The proposed framework converts the circuit under test into a directed graph, in which each node contains information such as device size, and each edge contains information about the connection of the circuit. The graph is then input into a trained gated graph sequence neural network~\cite{2015Gated}, and the cosine similarity of the two nodes is computed to detect symmetry after their final vectors are obtained. The disadvantage of this method is that edge computing overhead can become very large when facing circuits with complex connection relationships, thus facing challenges in promotion.

To address these issues, we propose a novel graph-based learning framework to automatically extract layout constraints of analog circuits.
Since the edges in netlist graph have different pin connections,
an edge-augmented graph attention network (EGAT) is proposed to extract the netlist information.
Compared with the traditional graph neural network, which focuses on node-level features, the proposed EGAT pays more attention to the interaction with node and edge features.
By utilizing graph neural networks, our framework can analyze existing circuits and learn general rules of symmetry constraints,
which in turn generalizes to new unseen circuits.
Our main contributions are summarized as follows:
\begin{itemize}
	\item
	An edge-augmented graph attention network (EGAT)-based learning framework with a new graph representation  is proposed to extract the netlist information effectively
    and measure the similarity of paired devices according to the resulting embeddings.
    %which is adopted to measure the similarity of pairs of devices.
    %which identifies common symmetric constraints in analog circuits by measuring the similarity of pairs of nodes.
    %A new graph representation for an analog circuit is built to model various analog circuit elements.
    %An efficient and concise method for embedding circuit features generated by device-level symmetric constraints is also designed.
    \item
    Suitable circuit features are designed to realize information interaction with nodes and edges.
    Meanwhile, an extra {4-dimensional} feature is introduced to distinguish differential pairs and current mirrors in analog circuits.
    %Meanwhile, an extra {4-dimensional} feature is introduced to distinguish differential pairs and current mirrors in analog circuits. %,common differential pairs from current mirrors in analog circuits.
    \item
    Several post-processing rules are developed to significantly reduce the false positive rate.
    \item
    Experimental results on typical analog circuit datasets demonstrate that the proposed graph learning framework achieves {$>99\%$ accuracy and {$>94\%$} F$_1$-score,
    significantly outperforming signal flow analysis (SFA) approach~\cite{xu2019magical}, the GNN-based extraction approach~\cite{gao2021layout}, and the unsupervised learning approach~\cite{chen2021universal} by a large margin.}
    %A significant improvement in accuracy was obtained compared with~\cite{2020S3det} and~\cite{2021Layout}.
\end{itemize}

The rest of the paper is organized as follows.
\Cref{sec:pre} gives the preliminaries and formulates the symmetric constraint extraction problem. %elaborates on the problem of symmetric constraint detection and then gives the problem formulation.
\Cref{sec:flow} describes the details of our proposed graph learning framework.
\Cref{sec:result} presents the experimental results, followed by the conclusions in \Cref{sec:concl}.

	\section{Preliminaries}
\label{sec:pre}

In this section,
the backgrounds of symmetric constraints in analog circuit layouts and graph neural networks are offered, and then we give the problem formulation.
%we will provide a specific introduction to symmetric constraints in analog circuit layouts, as well as a brief explanation of graph neural networks. Finally, we propose the problem of extracting symmetric constraints.

\subsection{Symmetric Constraints in Analog Circuit Layout}
\label{sec:map}

\begin{figure}[tb!]
	\centering
    \hspace{-.17in}
	\includegraphics[width=3.2in]{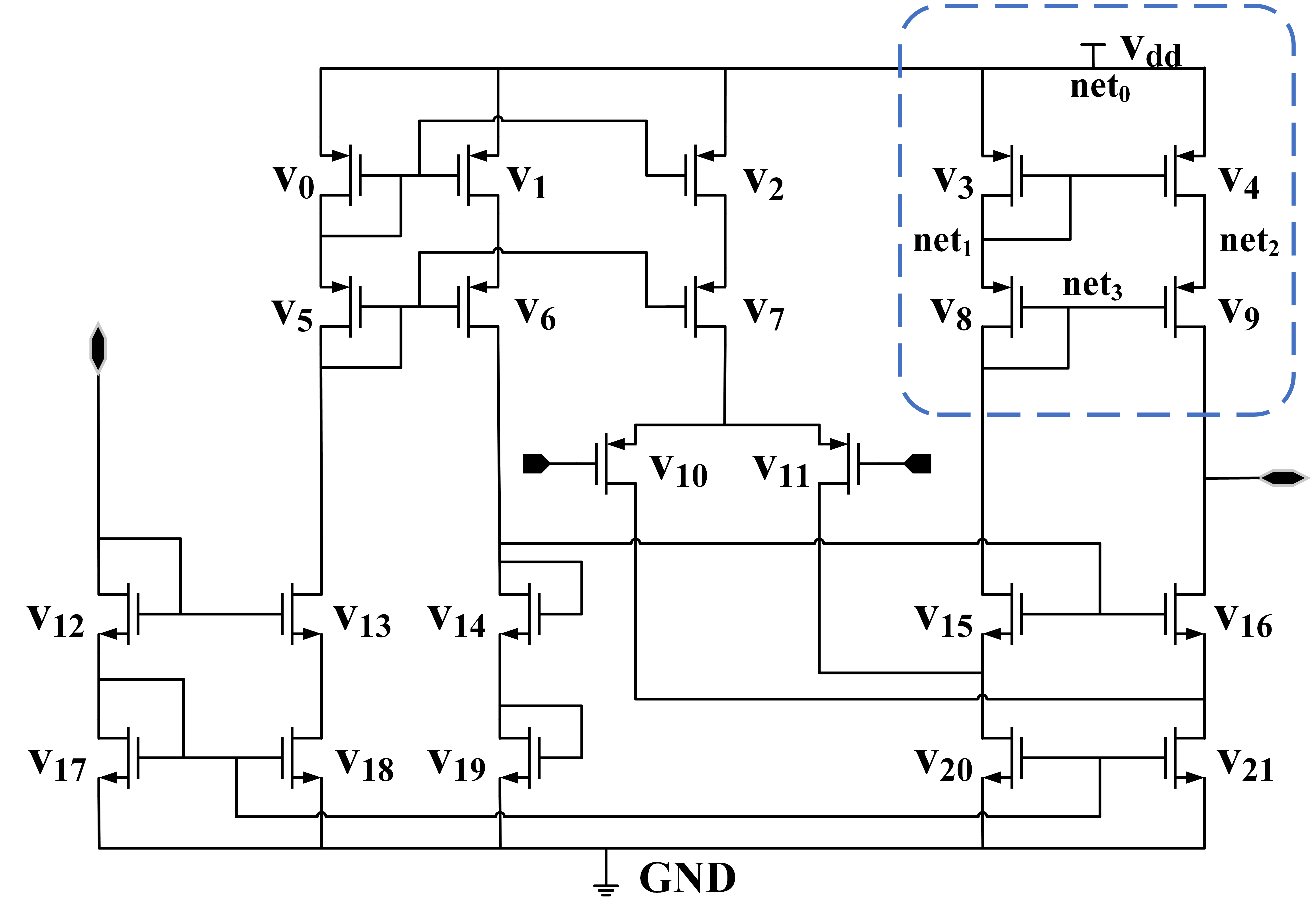}
	\caption{A typical OTA circuit.}
	\label{fig:2}
\end{figure}

In analog circuit systems,
layout symmetry constraints significantly impact circuit performance.
Based on the simple fact that the circuit netlist is a graph,
we represent the netlist of an analog circuit as a directed graph $G=(V, E)$.
$V$ denotes the nodes in the graph, representing circuit devices such as resistors, capacitors, transistors, etc.,
while $E$ describes the interconnection relationships among devices.
%In this work, we represent the analog circuit netlist as graph representations to extract device-level symmetry constraints. Specifically, we convert the network table file of an analog circuit into a graph $G=(V, E)$, where $V$ denotes the vertices in the graph, representing devices such as resistors, capacitors, and transistors in the circuit. $E$ represents the interconnection relationships between these devices.
%The symmetric constraint $(v_i, v_j)$ ($v_i,v_j\in V$) considered in this work are potential device pairs formed by pairwise combinations in the circuit.
For any pairwise combination, we need to detect the symmetrical device pair $(v_i, v_j)$ ($v_i,v_j\in V$), which should be placed symmetrically on the centerline of the layout.
%Our proposed graph learning framework is utilized to effectively extract these constraints.
\Cref{fig:2} depicts a typical operational transconductance amplifier (OTA) circuit consisting of multiple pairs of symmetrically matched devices,
i.e., $(v_0,v_1,v_2)$,
$(v_3,v_4)$, $(v_5,v_6,v_7)$, $(v_8,v_9)$, $(v_{10},v_{11})$,
$(v_{12},v_{13})$, $(v_{15},v_{16})$, and $(v_{17},v_{18},v_{20},v_{21})$.
%To illustrate, consider \Cref{fig:2}, which depicts a typical OTA circuit consisting of multiple pairs of symmetrically matched device pairs ($(M_0,M_1,M_2)$,$(M_3,M_4)$,$(M_5,M_6,M_7)$,$(M_8,M_9)$,$(M_{10},M_{11})$,
%$(M_{12},M_{13})$,$(M_{15},M_{16})$ and $(M_{17},M_{18},M_{20},M_{21})$). It should be noted that the same device can be matched with multiple others, which is a common occurrence in analog circuit design.

\subsection{Graph Neural Network}
\label{sec:gnn}

In recent years, traditional deep learning methods have achieved great success in extracting features from Euclidean space data.
However, there exists a large number of practical application scenarios,
where data are generated from non-Euclidean domains and are represented as graphs with complex relationships and interdependency between objects.
To handle the complexity of graph data,
the graph neural network is developed over the past few years.
For example, graph convolutional network (GCN)~\cite{kipf2016semi} promotes convolutional operations from traditional data, such as images, to graph data.
The main idea is to generate the representation of a node $v_i$ by aggregating its own features with its neighbors' features.
The GCN model is the basis of many complex graph neural network structures, including GraphSAGE network~\cite{hamilton2017inductive} and graph generative networks~\cite{simonovsky2018graphvae}.
But different from the GCN model, the GraphSAGE network adopts sampling to achieve a fixed number of neighbors for each node.
As a result, the efficiency of information interaction is improved.

\begin{figure}[tb!]
	\centering
	\includegraphics[width=.40\textwidth]{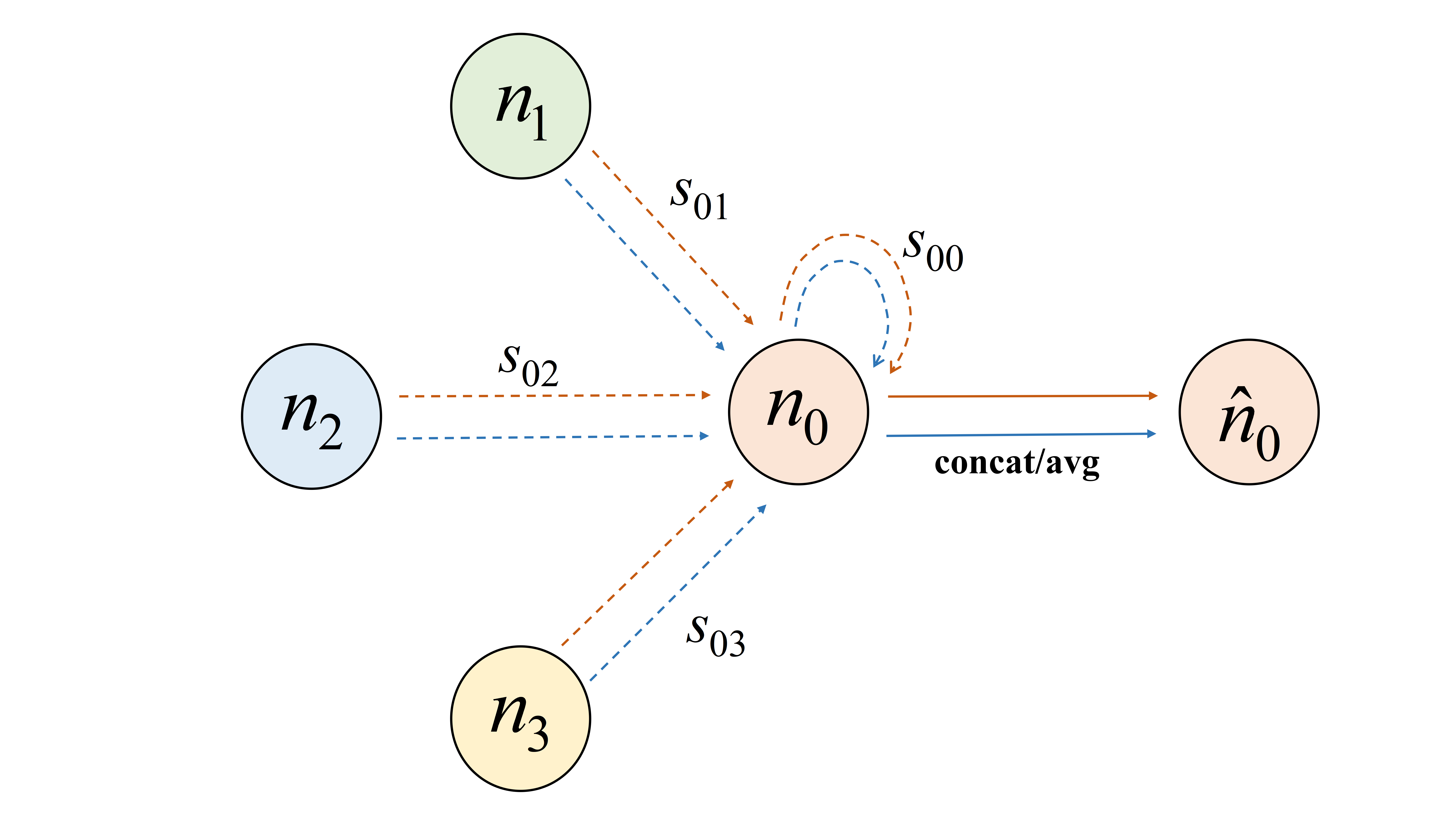}
	\caption{The illustration of the graph attention mechanism on node 0 with its neighborhood.
            %where $\vec{n}_i$ is the feature of node $i$ and $s_{0i}$ denotes the calculated attention score between node $i$ and node 0.
            %Different arrow colors and styles refer to independent attention computations (multi-head process).
            %Based on the calculated attention score $s_{0j}$, the feature of the neighboring node $j$ is aggregated in node 0.
            %Then, the aggregated features in each head are concatenated or averaged to achieve the final embedding $\hat{\vec{n}_0}$.
            }
	\label{fig:3}
\end{figure}

In many sequence-based tasks, attention mechanisms have become almost de facto standards,
which allow the model to focus on the most relevant parts of the input to make decisions.
When an attention mechanism is adopted to calculate a representation of a single sequence,
it is considered as a self-attention~\cite{vaswani2017attention}.
Based on the attention mechanism,
graph attention network (GAT)~\cite{velivckovic2017graph} is developed to learn the relative weights between two connected nodes.
Besides, in order to increase the expressive capability of the GAT model,
the multi-head attention is performed for the node embeddings.
\Cref{fig:3} depicts the simple aggregation process in the GAT network, where $\vec{n}_i$ is the feature of node $i$, and $s_{0i}$ denotes the calculated attention score between node $i$ and node 0.
Different arrow colors and styles refer to independent attention computations (multi-head process).
Based on the calculated attention score, the feature of the neighboring node $i$ is aggregated in node 0.
Then, the aggregated features in each head are concatenated or averaged to achieve the final embedding $\hat{\vec{n}_0}$. 
But the traditional GAT model only focuses on node-level features and cannot achieve feature interaction between nodes and edges.
To tackle the problem, an edge channel is introduced in~\cite{hussain2022global} to explicitly obtain the structural information of a graph.
Inspired by \cite{vaswani2017attention} and \cite{hussain2022global},
an edge-augmented graph attention network (EGAT)-based learning framework is proposed in this work to pay more attention to the interaction with node and edge features.
%which uses attention in the aggregation process,
%integrates the outputs of multiple models,
%and generates random walks targeting important targets. It is an efficient neural network framework. \Cref{fig:3} is a simple neural network aggregation form based on attention mechanism.
%Attention mechanism is a method that allows models to focus on relevant information rather than global information in data.
%It allows models to focus on specific parts that are meaningful to machine learning tasks. Attention mechanism is widely used in many fields, such as natural language processing, computer vision, etc.
%The self attention mechanism is a special attention mechanism that simultaneously considers all positions in a sequence and calculates the importance of each position. In the self attention mechanism, the model scores other positions in the sequence based on the current position, and weights and aggregates the information of all positions in the sequence based on these scores, thereby calculating the importance weight of each position.
%In graph neural networks, attention mechanisms can enhance information exchange between nodes, thereby improving model performance.
%Our graph learning model is inspired by~\cite{2017Attention} and~\cite{hussain2022global}, and effectively combines the information of edges in the graph with self attention mechanisms, thus achieving good results.

\subsection{Problem Formulation}
\label{sec:pf}

In this work, we formulate the symmetric constraint extraction as a binary classification problem.
%To evaluate the detection accuracy,
Several comprehensive measurements are defined to evaluate the detection quality,
including true positive rate (TPR), false positive rate (FPR), positive predictive value (PPV), accuracy (ACC), and F$_1$-score.
%In order to better solve the problem of symmetric constraint extraction in analog circuits, our model abstracts it into a simple binary classification problem, with matching and unmatching results. We also use several comprehensive measurement methods, including true positive rate (TPR), false positive rate (FPR), positive predictive value (PPV), accuracy (ACC), and $F_{1-score}$ of effective pairs, and define them as follows.

\begin{mydefinition}[TPR]
	TPR is the number of true positive results divided by the number of all positive results as:
    %measures how many samples that are actually positive are correctly judged as positive, and its formula is as follows in \Cref{00}.
\end{mydefinition}
\begin{equation}
    \textrm{TPR} = \frac{\textrm{TP}}{\textrm{TP}+\textrm{FN}}
	\label{tpr}
\end{equation}
where TP and FN are the number of the true positive and the false negative results.

\begin{mydefinition}[FPR]
	FPR is the number of false positive results divided by the number of all negative results as:
    %represents how many samples that are actually negative are erroneously judged as positive, and its formula is as follows in \Cref{01}.
\end{mydefinition}
\begin{equation}
    \textrm{FPR} = \frac{\textrm{FP}}{\textrm{FP}+\textrm{TN}}
	\label{fpr}
\end{equation}
where FP and TN denote the number of the false positive and the true negative results.

\begin{mydefinition}[PPV]
	PPV is the number of true positive results divided by the number of all predicted positive results as:
    %the proportion of samples that are predicted to be positive, but are actually positive. Its formula is as follows in \Cref{02}.
\end{mydefinition}
\begin{equation}
    \textrm{PPV} = \frac{\textrm{TP}}{\textrm{TP}+\textrm{FP}}
	\label{ppv}
\end{equation}

\begin{mydefinition}[ACC]
	ACC measures the proportion of correctly predicted samples to all samples, representing the overall accuracy of the prediction. %and is the most common evaluation method. Its formula is as follows in \Cref{03}.
\end{mydefinition}
\begin{equation}
    \textrm{ACC} = \frac{\textrm{TP}+\textrm{TN}}{\textrm{TP}+\textrm{FP}+\textrm{TN}+\textrm{FN}}
	\label{ACC}
\end{equation}

\begin{mydefinition}[F$_1$-score]
	F$_1$-score reflects the balance of the model in positive and negative case classification. %The formula is as follows in \Cref{04}.
\end{mydefinition}
\begin{equation}
    \textrm{F}_{1} = \frac{2\textrm{TP}}{2\textrm{TP}+\textrm{FP}+\textrm{FN}}
	\label{f1}
\end{equation}

%In these formulas, TP represents the true positive number of cases, representing the number of correctly identified true positive cases; FN represents the number of false negative cases, indicating the number of errors in identifying positive classes as negative classes. FP represents the number of false positive cases, indicating the number of negative classes incorrectly identified as positive classes; TN represents the true negative number, which represents the number of correctly identified true negative cases.

%In this work, our goal is to find as many symmetric constraint pairs and fewer asymmetric constraint pairs as possible in the given analog circuit netlist. With the evaluation metrics above, our problem is fomnulated as follows.
Based on the above metrics, we define the symmetric constraint extraction problem as follows.

\begin{myproblem}[Symmetric Constraint Extraction]
Given that various analog circuit netlists are labeled with symmetry pairs,
the goal is to train a graph learning model to detect the symmetry pairs from the new circuit netlist, yielding higher TPR, PPV, ACC, F$_1$-scores, and lower FPR.
%For a given analog circuit netlist N labeled with symmetric pairs, abstract it into a graph representation $G=(V, E)$ composed of v nodes and e edges. We believe that symmetric node pairs are highly similar, so we use the proposed graph learning model to identify node pairs with high similarity and strive to obtain higher TPR, PPV, ACC, F1 scores, and lower FPR as much as possible.
\end{myproblem}

\begin{figure*}[tb!]
	\centering
	\includegraphics[width=6.4in]{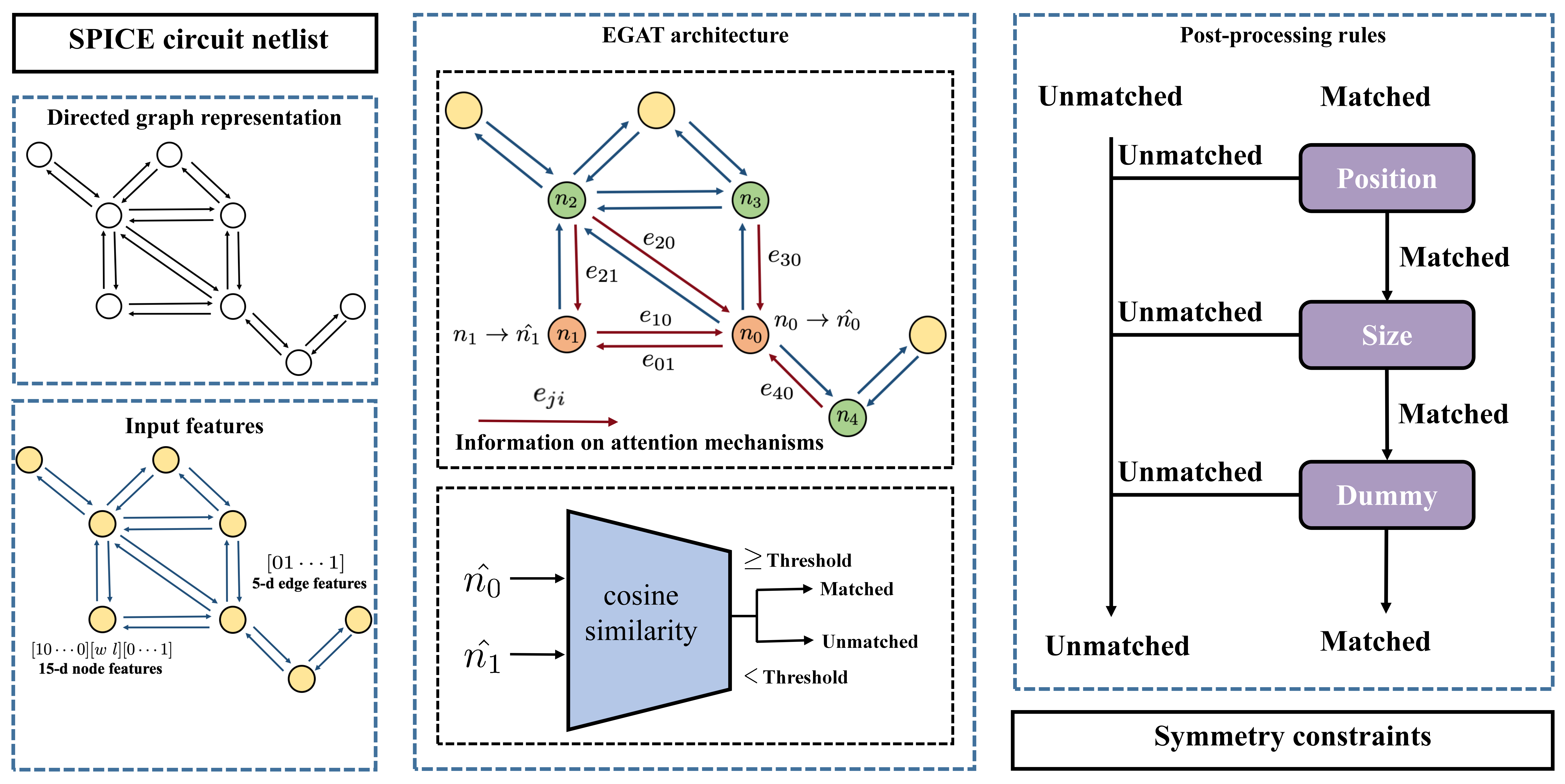}
	\caption{The diagram of the proposed EGAT-based symmetry constraint extraction framework.}
	\label{fig:4}
\end{figure*}

	\section{Proposed Graph Learning Framework}
\label{sec:flow}

To extract symmetric constraints in analog circuits,
we propose a graph learning-based framework as illustrated in \Cref{fig:4}, %based on our graph learning framework is shown in \Cref{fig:4}.
which consists of four main components, namely 1) directed graph representation, 2) input features, 3) edge-augmented graph attention network (EGAT) architecture, and 4) post-processing rules.
In the directed graph representation stage,
we convert the analog circuit netlist into a graph with bi-directional edges.
Then, we utilize partial information of the devices (type, size, etc.) as node features and embed the connection relationship of the devices as edge features,
which are then fed to the EGAT network architecture. %the second part of the graph neural network.
%In the learning phase of the graph neural network, we design a neural network EGAT that utilizes edge information and attention mechanisms to effectively distinguish symmetric matching pairs.
%The proposed EGAT network differs from the traditional attention method in GAT network in that edge features are introduced to calculate attention scores,
%which are further adopted to update both node and edge features.
Next, the proposed EGAT network effectively generates node embeddings,
which are adopted to predict the probabilities of symmetry constraint.
%As a result, the symmetric matching pairs can be distinguished effectively.
%The feature information from the first part serves as the input to the neural network.
Finally, in the post-processing stage,
several processing rules are designed to rectify the potential symmetry pair errors in EGAT recognition to improve accuracy. % improve the accuracy of the results by addressing the symmetric pairs of potential errors in neural network recognition.
The detailed techniques are described below.

\begin{figure}[tb!]
	\centering
	\includegraphics[width=2.8in]{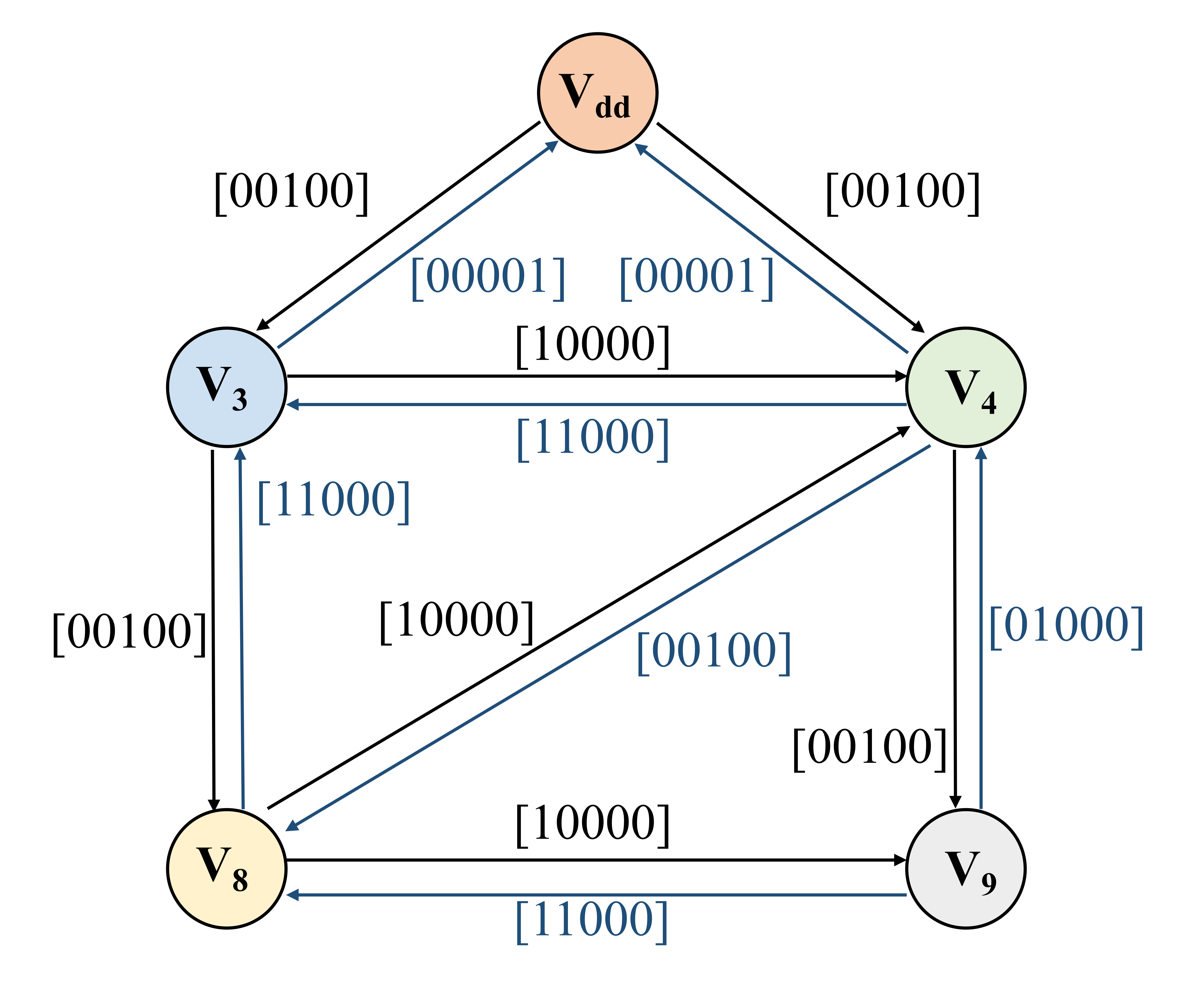}
	\caption{The directed graph representation for the partial circuit structure in \Cref{fig:2}. %the circuit structure in \Cref{fig:5}.
            %The edge connections are as follows: node %$v_3$ is connected to the gate port of node %$v_4$, the source port of node $v_8$, and the %power port $V_{dd}$;
            %node $v_4$ is connected to the gate and drain ports of node $v_3$, the source ports of nodes $v_8$ and $v_9$, and the power port $V_{dd}$;
            %node $v_8$ is connected to the gate and drain ports of node $v_3$, and the gate ports of nodes $v_4$ and $v_9$;
            %node $v_9$ is connected to the drain port of node $v_4$, and the gate and drain ports of node $v_8$.}
            %Graphic transformation display of edge information.
            %Among them, $N_1$ is connected to the gate port of $N_2$ and the source port of $N_3$. $N_2$ is connected to the gate and drain port of $N_1$, $N_3$'s source port and $N_4$'s source port. $N_3$ is connected to the gate and drain port of $N_1$, $N_2$'s gate port and $N_4$'s gate port. $N_4$ is connected to the drain port of $N_2$,$N_3$'s gate and drain port, respectively.
            }
	\label{fig:6}
\end{figure}

\subsection{Directed Graph Representation}
\label{sec:dgr}
In this section,
we first explain how to represent a SPICE circuit netlist as a directed graph. %, and then present the feature embedding process.
%\Cref{fig:5} is a partial structure of the OTA circuit in \Cref{fig:2} (blue dashed lines), where $v_3$, $v_4$, $v_8$, $v_9$ are the devices and $Vdd$ is the IO port.
%In the figure, we represent them as nodes $N_i (i \in (1,2,3,4))$, and in addition to the components, we also consider the IO ports connected to the circuit as vertices, which are also added to the figure.
For analog circuits, devices and IO ports are nodes of the graph,
while the nets connecting devices or IO ports are recognized as edges. %and edges are the connection
%\Cref{fig:5} is a partial structure of the OTA circuit in \Cref{fig:2} (blue dashed lines),
For a partial structure of the OTA circuit in \Cref{fig:2} (blue dashed lines),
$v_3$, $v_4$, $v_8$, and $v_9$ are the devices, $V_{dd}$ is the power IO port, and $net_0$, $net_1$, $net_2$, and $net_3$ refer to the nets.
%The networks $net_0$, $net_1$, $net_2$, and $net_3$ that connect various devices in the circuit are represented as edges $E_i (i \in (1,2,3,4))$ in the diagram.
Since different connection directions between nodes indicate varying connection information,
bi-directional edges between any two nodes are constructed in the graph.
Thus, the number of edges depends on the number of nodes and the circuit's connectivity.
The corresponding directed graph representation is depicted in \Cref{fig:6}.
As illustrated in the graph, node $v_3$ has bi-directional edges with nodes $v_4$, $v_8$, and $v_{dd}$, respectively.

%As the connections between nodes are mutual rather than unidirectional, there are two opposing lines between any two nodes in our diagram representation.

%\begin{figure}[tb!]
%\label{picure2}
%	\centering
%	\includegraphics[width=2.7in]{figs/fig5}
%	\caption{Partial structure of the OTA circuit in \Cref{fig:2}.}
%	\label{fig:5}
%\end{figure}

\subsection{Input Features}
\label{sec:if}
%We use the following method for graph feature embedding.
After the directed graph construction,
we define an initial feature vector for each node.
Since the size and type of the node are important factors for effectively identifying symmetric constraints,
%we convert the node type (i.e., NMOS, PMOS, NPN, PNP, diode, passive, IO) into a unique one-hot representation of a 7-dimensional vector.
we convert the node type (i.e., NMOS, PMOS, NPN, PNP, diode, resistance, capacitance, inductance, IO) into a unique one-hot representation of a 9-dimensional vector.
%During the analog circuit design,
%designers need to specify the adopted Process Design Kit (PDK) as well as the cell libraries for each device.
{Note that matched devices should be indeed identical,
not just the same type or from the same library.}
Moreover,
the size information of a device is represented as a 2-dimensional vector, where length and width serve as one dimension, respectively. % after converting to a unified unit of micrometers
Besides, the influence of fingers is also taken into account.
{In the circuit design process, when the width of a device is set to $w$ (unit width), and the number of fingers is $n_f$,
the device in the schematic will have a width of $w$ (unit width), but in the SPICE netlist, the device width obtained is $w\cdot n_f$ (total width).
Therefore, we need to divide the number of fingers $n_f$ to get the device's unit width $w$.}
%\textcolor{blue}{
%We specify the device's width and length as $w$ and $l$. The number of fingers is denoted by $n_f$, and $W$ represents the total width.
%In the layout, multipliers represent the number of parallel device, denoted as $M$.
%Regardless of multipliers, symmetrical devices should be symmetrically placed in the layout.
%Therefore, multipliers does not affect the symmetry detection, so we do not consider the impact of $M$. But symmetry is related to the unit width, and in the SPICE netlist, $W = w\cdot n_f$, so we need to divide $W$ by $n_f$ to obtain $w$, while $l$ remains unchanged.}
For instance, if a device has a length and width of 10$\mu$m with a finger value of 2, the length remains unchanged, while the width needs to be divided by the finger value.
As a result, the corresponding 2-dimensional feature is [10,5].
To facilitate neural network training, the 2-dimensional size vector of device nodes is normalized with values between [0,1],
while the IO node's size is set to a default value of -1.
%Note that, the IO node's size is set to a default value of -1, while the sizes of other nodes are normalized with values between [0,1].
%and normalize the size information (length and width) of the node into a 2-dimensional vector with values between [0,1].
%Note that the IO node's size is set to a default value of -1.
%For the IO node, the size is set to a default value of -1.
%In addition, we found in the experiment that due to the widespread presence of differential pairs and current mirrors in analog circuits, significant differences in gate connection relationships can lead to some misjudgments in the neural network.

Due to many differential pairs and current mirrors exist in the analog circuit,
and the main difference between them is the connection relation of the gates,
%in order to make the network perceive the connection difference,
an extra {4-dimensional} feature is introduced to make the network perceive the connection difference.
%\Cref{alg:2} gives a high level description of the algorithm to determine the 4-dimensional gate connectivity feature.
The detailed description of this 4-dimensional feature is as follows.
If the device is not a MOSFET, the feature is configured as [1,0,0,0].
For the PMOS (NMOS) device,
when its gate port is connected to the gate port of another PMOS (NMOS) device,
the feature is set to [0,1,0,0].
But if the above condition does not hold and the gate port of the device is linked to an IO,
its feature is adjusted to [0,0,1,0].
In all other cases, the feature is designated as [0,0,0,1].
Following the specified order above ensures a prioritized assignment,
preventing potential ambiguities.
For instance,
%The prioritization of assignments follows the specified order above, thus avoiding any potential ambiguities.
in \Cref{fig:2},
although the gate port of NMOS device $v_{12}$ is connected to the IO port,
its gate port is also linked to the gate port of NMOS device $v_{13}$,
connections to the gate ports of the MOSFET have higher priority.
So the corresponding feature of $v_{12}$ is set to [0,1,0,0].
%they connected to each other and are both NMOS devices, thus the feature is ste to [0,1,0,0], adhering to the established assignment sequence.}
%The 1-dimensional feature is set to 1 when the gate port of the node is connected to the IO port and 0 otherwise.
%To address this, we introduced 1-dimensional additional vectors to nodes to assist in the training of the neural network.
%We set this 1-dimensional vector to 1 when the gate port of the node is connected to the IO port, and 0 otherwise.
%Algorithm \ref{alg:1} provides a detailed description of this process.
After obtaining the vector representations of three parts,
we concatenate them into a {15-dimensional} vector as the final node feature.

In addition, a 5-dimensional multi-hot vector is encoded for edges to represent the connection information.
The first four dimensions indicate the connections to the gate port, drain port, source port of MOS, and passive device,
while the last dimension denotes other possible pin connections (NPN, PNP, etc.).
\Cref{fig:6} illustrates the construction process of edge features.  %of the circuits structure. %in \Ccref{fig:5}.
The edge connections are as follows: node $v_3$ is connected to the gate port of node $v_4$, the source port of node $v_8$, and the power port $V_{dd}$;
node $v_4$ is connected to the gate and drain ports of node $v_3$, the source ports of nodes $v_8$ and $v_9$, and the power port $V_{dd}$;
node $v_8$ is connected to the gate and drain ports of node $v_3$, and the gate ports of nodes $v_4$ and $v_9$;
node $v_9$ is connected to the drain port of node $v_4$, and the gate and drain ports of node $v_8$.
Therefore, based on the constructed graph connections, the feature for the edge from from $v_3$ pointing to $v_4$ is [1,0,0,0,0], while the feature for the edge from $v_4$ pointing to $v_3$ is [1,1,0,0,0].
Other edge features are also represented in the graph, with the colors of edges matching their respective features to make the illustration more clear.
%For example, node $v_3$ connects to the gate port of node $v_4$, and thus the feature for the edge from node $v_3$ pointing to $v_4$ is [1,0,0,0,0].
%Similarly, node $v_4$ connects to the gate and drain ports of node $v_3$, so the feature for the edge from node $v_4$ pointing to $v_3$ is [1,1,0,0,0].
%representing the pin connections of gate, drain, source, and passive respectively, and the last dimension represents other possible pin connections.
%When there is a connection relationship between node $v_i$ and a pin of node $v_j$,
%the vector corresponding to the directed edge e of node $v_i$ pointing to $v_j$ is represented by 1, otherwise it is 0.
%into a graphical representation.
\Cref{table:1} summarizes the initial features and dimensions of nodes and edges, %, which are used for graph learning.
which are concatenated into vector representations and then passed to the downstream graph network.

\subsection{EGAT Architecture}
\label{sec:gnnl}

\begin{figure}[t]
	\centering
	\includegraphics[width=.35\textwidth]{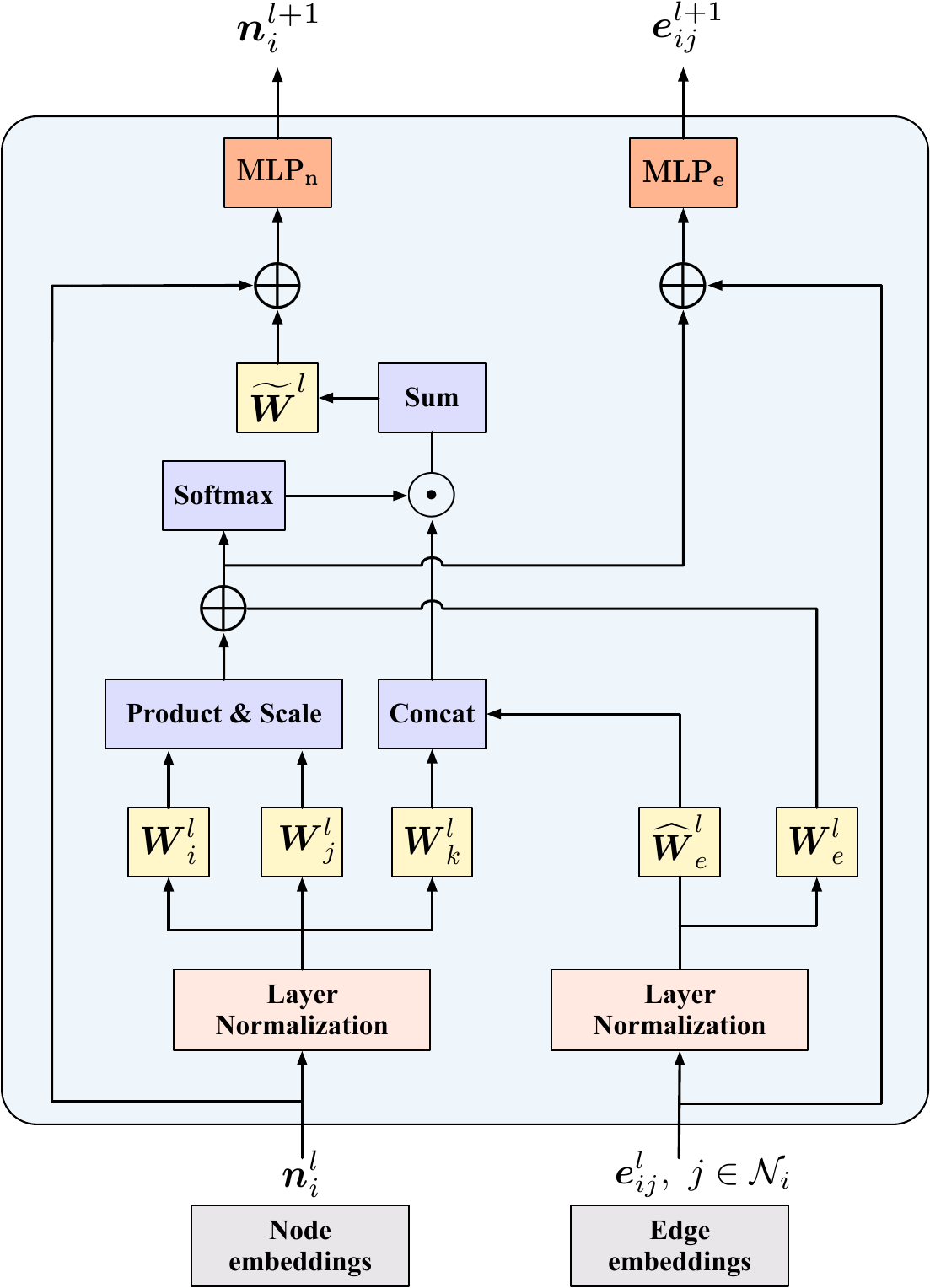}
	\caption{The architecture of the EGAT network.}
	\label{fig:arch}
\end{figure}

%To enhance the generalizability of our proposed framework,
%we train an inductive Graph Neural Network model, named EGAT.
%Unlike transductive models,
%which establish a function between input and output data,
%inductive models can induce patterns from training data and make predictions.
%As such, our model can recognize symmetric pairs in previously unseen circuit diagrams.
Based on the input node and edge features,
the developed EGAT network further extracts the embeddings of nodes and edges.
In order to increase the expressive capability of the model,
an edge-augmented attention mechanism is performed for each node.
Each layer $l$ in the EGAT consists of four calculation processes:
layer normalization, attention score calculation, node embedding update, and edge embedding update, as shown in~\Cref{fig:arch}.

Layer Normalization (LN)~\cite{ba2016layer} plays a crucial role in standardizing inputs,
avoiding the exploding or vanishing gradients, and accelerating the training process.
The calculation process of LN is as follows.
%reducing the variance of training gradients, accelerating the training process, and aiding the model in learning more robust representations.
%LN enables our model to better adapt to various input conditions, thereby improving its generalization ability. The calculation equation for LN is as follows in \Cref{1}:
\begin{equation}
	\begin{aligned}
		&\mu^{l} = \frac{1}{{n}^{l}} \sum_{i=1}^{{n}^{l}} {x_{i}}^{l},\ (\sigma^{l})^{2} = \frac{1}{{n}^{l}} \sum_{i=1}^{{n}^{l}}  ({x_{i}}^{l} - \mu^{l})^2,\\
        &\vec{\hat{x}}^{l} = \frac{\vec{x}^{l} - \mu^{l}}{\sqrt{(\sigma^{l})^{2} + \epsilon}},\ \mathrm{LN}_{\vec{\alpha},\vec{\beta}}(\vec{x}^{l}) = \vec{\alpha} \odot \vec{\hat{x}}^{l} + \vec{\beta},
        %\vec{x}^{l+1} = \vec{\gamma} \odot \vec{\hat{x}}^{1} + \vec{\beta},
         %\Leftarrow LN_{\vec{\gamma},\vec{\beta}}(\vec{z}^{1})
	\end{aligned}
	\label{1}
\end{equation}
where $\vec{x}^l$ is the input of the $l$-th layer, and $n^l$ refers to the number of neurons in the $l$-th layer.
%$n^{(l)}$ represents the number of neurons in l-th layer,
$\mu^{l}$ and $\sigma^{l}$ represent the mean and the standard deviation of the input data, respectively.
Besides, $\epsilon$ is a small constant, while $\vec{\alpha}$ and $\vec{\beta}$ are learnable parameters with the same dimension as $\vec{x}^l$.
$\odot$ is the Hadamard product.
In this work, we first perform the LN on the initial features as follows.
%$\sigma^{{(l)}^{2}}$ represents the standard deviation of the input data, $\epsilon$ is a small constant used to avoid situations where the denominator is zero, $\vec{\gamma}$ and $\vec{\beta}$ represent parameter vectors for scaling and translation, with the same dimension as $\vec{z}^{(l)}$. \Cref{2} performs Layer Normalization on the input node features and edge features respectively:
%\begin{equation}
%	\vec{\hat{n}}_i^l = \mathrm{LN}_{\vec{\alpha},\vec{\beta}}(\vec{n}_i^{l}),\ \
%        %{\vec{NL}_{i}}^{(l)} = LN_{\vec{\gamma,\beta}}({\vec{N}_{i}}^{(l)}) \\
%    \vec{\hat{e}}_{ij}^l = \mathrm{LN}_{\vec{\alpha},\vec{\beta}}(\vec{e}_{ij}^l),
%	\label{2}
%\end{equation}
\begin{equation}
	\vec{\hat{n}}^l = \mathrm{LN}_{\vec{\alpha},\vec{\beta}}(\vec{n}^{l}),\ \
        %{\vec{NL}_{i}}^{(l)} = LN_{\vec{\gamma,\beta}}({\vec{N}_{i}}^{(l)}) \\
    \vec{\hat{e}}^l = \mathrm{LN}_{\vec{\alpha},\vec{\beta}}(\vec{e}^l),
	\label{2}
\end{equation}
where $\vec{n}^{l}$ and $\vec{e}^l$ are the embeddings of all nodes and edges in the $l$-th layer.
As the $\mathrm{LN}$ technique enables the model to adapt to various inputs, the model can generalize to new unseen netlists.
%feature vector of the l-th layer, and ${\vec{E}_{i}}^{(l)}$ represents the edge feature vector of the l-th layer.

\begin{table}
\centering
\renewcommand{\arraystretch}{1.2}
\caption{Designed features of nodes and edges in the directed graph representing the analog circuit netlist.}
\resizebox{8.5cm}{!}{
    \begin{tabular}{c|c|c}
    \toprule
    Type & Feature Description & Dimension \\
    \midrule
    \multirow{3}{*}{Node} & One-hot representation for device type      & {9}  \\
                          & Length and width of device                  & 2  \\
                          & Connection relationship of gate port
                          & {4}  \\
    \midrule
    Edge                  & Connection information between pins         & 5  \\
    \bottomrule
    \end{tabular}
    }
    \label{table:1}
\end{table}

Then the multi-head attention mechanism is performed over the learned node representations produced by $\mathrm{LN}$.
The attention score of node $v_i$ with its neighbouring node $v_j$ are computed as follows.
\begin{equation}
\label{attens}
\begin{aligned}
%&\vec{a}_{ij}^{l}=\vec{W}_i^l\vec{\hat{n}}_i^l\odot %{\vec{W}_j^l\vec{\hat{n}}_j^l}+\vec{W}_e^l\vec{\hat{e}}_{ij}^l, \\
&\vec{a}_{ij}^{l}=\mathrm{reshape}((\vec{W}_i^l\hat{\vec{n}}_i^l)\odot ({\vec{W}_j^l}\hat{\vec{n}}_j^l))+\vec{W}_e^l\hat{\vec{e}}_{ij}^l, \\
%&s_{ij}^{l}=\mathrm{softmax}(\vec{a}_{ij}^{l}),%=\frac{\mathrm{exp}{(\vec{a}_{ij}^{l})}}{\sum\limits_{k\in \mathcal{N}_i}{\mathrm{exp}(\vec{a}_{ik}^{l})}},
&s_{ij}^{l}=\frac{\mathrm{exp}{(\vec{b}^{l^{\top}}\vec{a}_{ij}^{l})}}{\sum\limits_{j\in \mathcal{N}_i}{\mathrm{exp}(\vec{b}^{l^{\top}}\vec{a}_{ij}^{l})}},
\end{aligned}
\end{equation}
where $\hat{\vec{n}}_i^l\in{\mathbb{R}^{d_n}}$, $\hat{\vec{n}}_j^l\in{\mathbb{R}^{d_n}}$ and $\hat{\vec{e}}_{ij}^l\in{\mathbb{R}^{d_e}}$ are the embeddings of node $v_i$, node $v_j$ and the corresponding edge. %in the $l$-th layer,  respectively.
$\vec{W}_i^l\in{\mathbb{R}^{d_n\times d_n}}$, $\vec{W}_j^l\in{\mathbb{R}^{d_n\times d_n}}$, $\vec{W}_e^l\in{\mathbb{R}^{d_e\times d_e}}$ and $\vec{b}^l\in{\mathbb{R}^{d_e}}$ denote learning parameters.
$\mathrm{reshape}(\cdot)$ refers to the reshape operations.
%and $\mathrm{softmax}(\cdot)$ function is adopted to normalize the correlation result.
Besides, $\mathcal{N}_i$ represents the neighborhoods of node $v_i$ in the graph.
In reality, due to the multi-head attention mechanism, %to allow for parallel computation of multiple head,
the above calculation is decomposed in multiple heads. %are projected into multiple linear projections in a column-parallel manner.
%in order to simultaneously calculate multiple attention heads, matrices WQ, WK and WV are divided into multiple
The calculated attention score $s_{ij}^l$ indicates the importance of node $v_j$'s embedding to node $v_i$.
%Since the edge features influence the calculation of attention scores,
%the information interactions between nodes and edges are realized.
%while letting edges influence the calculation of attention scores,

On the basis of the attention scores, the node embeddings are updated by the weighted average pooling operation.
To further enhance the representation ability of the model,
we use an MLP to carry out the non-linear transformation of the pooled node embeddings.
Besides, the residual connection is performed to facilitate training networks~\cite{he2016deep}.
The final node representation is updated as:
\begin{equation}
\label{nupdat}
\vec{n}_i^{l+1}=\mathrm{MLP_n}(\vec{n}_i^{l}+\widetilde{\vec{W}}^l\sum_{j\in \mathcal{N}_i}{s_{ij}^{l}\cdot \mathrm{concat}(\vec{W}_k^l\hat{\vec{n}}_j^{l}, \widehat{\vec{W}}_e^l\hat{\vec{e}}_{ij}^l)}), \\
\end{equation}
where $\vec{W}_k^l\in{\mathbb{R}^{d_n\times d_n}}$, $\widehat{\vec{W}}_e^l\in{\mathbb{R}^{d_e\times d_e}}$ and $\widetilde{\vec{W}}^l\in{\mathbb{R}^{(d_n+d_e)\times d_n}}$ refer to learning weight vectors.
$\mathrm{concat}(\cdot)$ represents the concatenation operation.
$\mathrm{MLP_n}(\cdot)$ denotes a non-linear transformation with input dimension $d_n$ and output dimension $d_n$.

Similarly, the edge representation is also updated with a residual connection as follows:
\begin{equation}
\label{eupdat}
\vec{e}_{ij}^{l+1}=\mathrm{MLP_e}(\vec{e}_{ij}^l+\vec{a}_{ij}^{l}),
\end{equation}
where $\mathrm{MLP_e}(\cdot)$ refers to another non-linear transformation with input dimension $d_e$ and output dimension $d_e$,
and $\vec{a}_{ij}^l$ is the intermediate feature calculated in \Cref{attens}.

As demonstrated above,
the proposed EGAT network differs from the traditional attention method in GAT network~\cite{velivckovic2017graph} in that edge features are introduced to calculate attention scores,
which are further adopted to update both node and edge features.
As a result, the information interactions between nodes and edges are realized.
Since the edge features reflect the pin connectivity between nodes,
by incorporating the edge feature into the calculation of the attention score,
the symmetric matching pairs can be distinguished effectively.
Therefore, EGAT makes full use of the characteristics of the netlist graph for the symmetric constraint extraction in analog circuits.

Based on the node embeddings generated by the EGAT architecture,
we should provide a predicted similarity score for each candidate symmetry pair $(v_i, v_j)$ ($v_i,v_j\in V$).
As the symmetric constraint extraction is formulated as a binary classification problem,
labels are annotated in advance, where the symmetric and asymmetric node pairs are labeled as 1 and -1 respectively. %the asymmetric pair is -1.
To match with the labels,
the $\mathrm{cosine}$ similarity function is adopted to predict the probabilities of symmetry constraint.
%In order to better calculate the similarity of symmetric pairs recognized by neural networks, we adopted the widely used cosine similarity function~\cite{1975The}. The range of cosine similarity values is $[-1, 1]$. When two vectors are exactly the same, the cosine similarity is 1; When two vectors have completely opposite directions, the cosine similarity is -1. In the labels we annotated in advance, the symmetric node pair is 1, and the asymmetric node pair is -1, which is completely consistent. The formula for the cosine similarity function is as follows in \Cref{7}:
\begin{equation}
    \mathrm{similarity} = \frac{\vec{n}_i \cdot \vec{n}_j}{|\vec{n}_j| |\vec{n}_j|},
	\label{7}
\end{equation}
where $\vec{n}_i$ and $\vec{n}_j$ are the embeddings of nodes $v_i$ and $v_j$ produced by the EGAT.

Due to the utilization of the $\mathrm{cosine}$ similarity function,
the binary cross-entropy loss~\cite{bishop2006pattern} originally used to infer results in the range [0,1] is no longer valid.
%the original binary cross-entropy loss~\cite{bishop2006pattern} infer results within the range of $[0,1]$ is no longer valid.
To accommodate this,
the logistics loss function is employed for the output label with 1 and -1 as:  %based on the output label ${-1, +1}$.
%The expression of this function is as follows in \Cref{8}:
\begin{equation}
    loss = \log{(1+e^{-gs})},
	\label{8}
\end{equation}
where $gs$ is the product of the ground truth and the similarity score,
which reflects the accuracy and the confidence level of prediction.
That is, when $gs\geq0$, the prediction is correct, and the larger the value, the higher the confidence level.
On the contrary, when $gs<0$, the prediction is incorrect, and the smaller the value, the lower the confidence level.
%The symbol of $ys$ reflects the accuracy of the prediction, and the numerical value of $ys$ also reflects the confidence level of the prediction. When $ys\ge0$, the prediction is correct, and the larger the value, the higher the confidence level. On the contrary, when $ys\small0$, the prediction is incorrect, and the smaller the value, the lower the confidence level.

\begin{algorithm}[tb!]
\begin{flushleft}
     \textbf{Input}: A directed graph $G$ for a circuit.\\
     \textbf{Output}: The position $p_n$ of each device.
    \end{flushleft}
\begin{algorithmic}[1]
\Statex \Comment{Assign weights for edges}
\For{each edge $e$ in $G$}
 \State Get two devices connected by $e$ as $d_i$ and $d_j$;
 \If{$d_i$ and $d_j$ are passive devices}
    \State Set the weight of $e$ to 0;
 \ElsIf{$d_i$ or $d_j$ is a passive device}
    \State Set the weight of $e$ to 0.5;
 \Else
    \State Set the weight of $e$ to 1;
 \EndIf
\EndFor
\Statex \Comment{Find the device position}
\For{each node $n$ in $G$}
    \If{$n$ is a PMOS}
        \State Calculate the shortest path from power node to $n$;
        \State $p_n$ = the weighted length of the path;
    \ElsIf{$n$ is a NMOS}
        \State Calculate the shortest path from GND node to $n$;
        \State $p_n$ = the weighted length of the path;
    \Else
        \State $p_n$ = $0$.
    \EndIf
\EndFor
\end{algorithmic}
\caption{The Pseudo-code of Device Position Calculation}
\label{alg:1}
\end{algorithm}

\subsection{Post-Processing Rules}
\label{sec:pp}

Although the graph learning framework can annotate the symmetry constraints, %we have achieved satisfactory performance through graph-based learning,
a case still exists where asymmetric pairs are identified as symmetry constraints.
Thus, we further develop several post-processing rules to reduce the false positive rate (FPR).
Note that the developed post-processing rules are only employed to correct the output of the model during the testing (inference) phase.
%reduce asymmetric pairs that may be identified due to neural network errors and to enhance overall accuracy, post-processing of the model's output is necessary.

%Firstly,
%given that in most circuits, symmetrical matching pairs usually share a common net,
%a post-processing rule is designed to eliminate wrongly identified symmetric pairs without shared nets.
%given that the majority of symmetrical matching pairs in the circuit share a common net,}
%We discovered that despite training the neural network with features that include the type and size of devices, there are still some device pairs with varying sizes that are incorrectly recognized. Therefore, we apply post-processing techniques to the output results to eliminate these wrongly identified symmetric pairs. This approach is entirely reasonable since symmetric matching pairs must have the same size.

Firstly,
we observe that symmetric device pairs tend to have identical device positions, as defined below.
%To define the global position of a device, we employ the following approach:
\begin{mydefinition}[Device Position]
	For each PMOS device,
    we find the shortest path from the power node to the device.
    Conversely,
    for each NMOS device, the shortest path from the GND node to the device is calculated.
    The weighted length of the shortest path refers to the device position.
\end{mydefinition}

\begin{figure}[t]
	\centering
	\includegraphics[width=.46\textwidth]{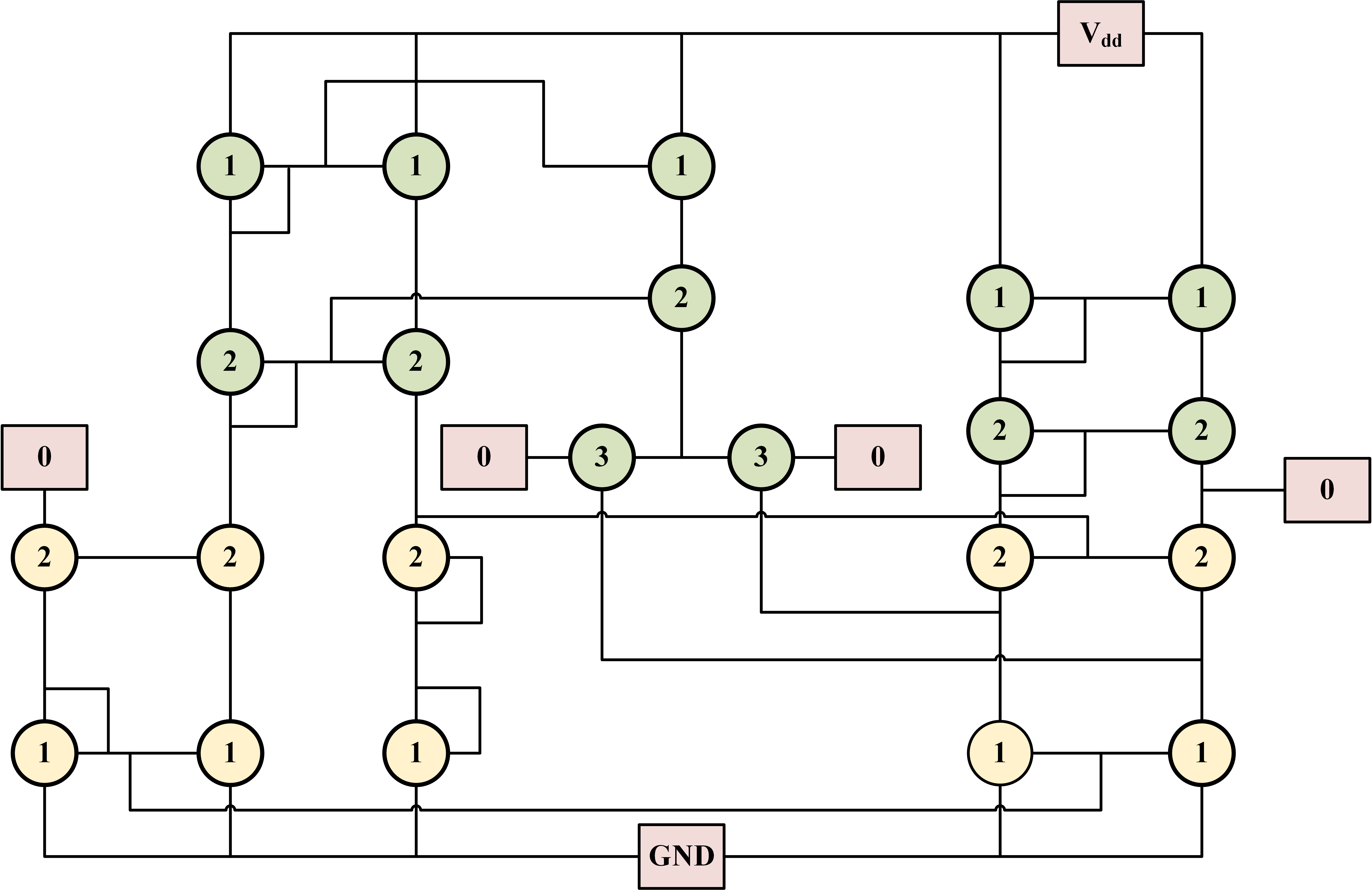}
	\caption{The illustration of device positions, where the green and yellow circles represent PMOS and NMOS devices and the pink rectangle denotes IO ports.}
	\label{fig:7}
\end{figure}

\Cref{alg:1} depicts the process of determining the positions of the devices in an analog circuit.
\Cref{fig:7} shows the device positions of the OTA circuit in \Cref{fig:2}.
It can be seen that symmetric devices exhibit the same device positions.

The second rule is that symmetric pairs must have the same size.
Although the designed features of nodes in the neural network contain the type and size information,
some device pairs with varying sizes are incorrectly recognized as symmetric pairs after the EGAT recognition. 
For instance, due to functional requirements in circuit design, devices of the same type but different flavors (models) usually exhibit differences in size. 
Thus, if the mismatch occurs in the neural network's output,
this post-processing rule will eliminate it.
Taking the NMOS as an example, mismatching between low-Vth (threshold voltage) NMOS and regular-Vth NMOS can be effectively removed by the rule.
%\textcolor{blue}{Due to functional requirements in circuit design, devices of the same type but different flavors (models) often exhibit differences in size. If mismatch occurs in the output of the neural network, this post-processing rule will also remove it. For example, in NMOS type, matching between nmos-low-VT and nmos-regular-VT would be effectively removed by this rule.}
%For example, due to functional requirements in circuit design, different device flavors usually have size differences.
%Thus, if the network model outputs mismatches due to device flavors,
%we can eliminate them by this post-processing rule.
%Thus, this post-processing rule can rectify the potential symmetry pair errors.
%based on the fact that devices lacking a shared net connection cannot be matched,

%At the same time, considering that different analog circuit engineers have different habits when designing circuits, some engineers may add dummies to the circuit, which are used to fill in missing parts of the network or separate certain parts of the network. They do not contain actual functions and are designed purely for the performance, stability, or layout of the circuit. These are also unnecessary constraints that we need to eliminate.

%Finally,
%dummies are generally added to reduce the mismatch between transistor pairs or fill in missing parts of the layout.
%Thus, dummies contain no actual functionality and are designed purely for the performance and stability of the circuit layout,
%which are also unnecessary constraints that we need to eliminate.

In analog circuit schematics,
dummies refer to virtual devices without any actual electrical functionality in the circuit design.
Instead, they are introduced to meet the requirements of the layout or to be in conjunction with the corresponding devices in the layout.
Thus, if a matching pair contains dummy devices,
we will eliminate it in the post-processing.
%Thus, in this work, we do not identify matches for dummy devices.}

%
%    \For {each $pin_{i}$ in $net$}
%        \For {each $pin_{j}(i<j)$ in $net$}
%            \State $device_i$ = get $p_i$'s device
%            \State $device_j$ = get $p_j$'s device
%            \If{$device_i$ is not $device_j$}
%                \If{no edge between $device_i$ and $device_j$}
%                    \If {$device_i$ and $device_j$ is passive}
%                        \State Add a bidirectional edge with $edge_w$ of 0
%                    \EndIf
%                    \If {$device_i$ or $device_j$ is passive}
%                        \State Add a bidirectional edge with $edge_w$ of 0.5
%                    \EndIf
%                    \If {$device_i$ and $device_j$ is not passive}
%                        \State Add a bidirectional edge with $edge_w$ of 1
%                    \EndIf
%                \EndIf
%            \EndIf
%        \EndFor
%    \EndFor

	\section{Experimental Results}
\label{sec:result}

\subsection{Simulation Setup}
\label{sec:ss}

The proposed graph learning framework is implemented in \texttt{Python} with the \texttt{Pytorch} library~\cite{DL-NIPSW2017-PyTorch},
and we execute it on a Linux server with the Intel Xeon Gold 6254 CPU and Nvidia Tesla V100 GPU.
The proposed EGAT architecture includes a three-layer network with five attention heads.
The threshold for the similarity score calculated in \Cref{7} is set to 0.6.
That is, if the similarity score of a valid pair is greater than 0.6,
the pair will be annotated as a symmetry constraint.
Unlike the most commonly used threshold of 0.9 or higher,
the low threshold ensures that more symmetric pairs are correctly identified, despite potentially resulting in a higher false alarm rate.
But the developed post-processing can effectively eliminate these misidentified pairs.
%During the training, each batch consists of 256 valid pairs.
In the experiment, we perform 500 training epoches, with a batch size of 256 for each epoch.
The initial learning rate is set to 0.002.
Besides, the adaptive moment estimation (Adam) optimizer~\cite{kingma2014adam} is adopted to train the network models.
We have open-sourced the codes on GitHub $\footnote{\url{https://github.com/wanglijie1999/Analog-EGAT-SymExtract}}$.
%All datasets are also available in the repository.
%In addition, due to the circuits being part of a designed system with IP sensitivity,
%only a portion of the symmetric labels in datasets is released for running the code, which can be found in the ``example'' folder at the link.}
%We will release the detailed codes soon.
%with each training epoch consisting of 256 valid pairs per batch.

\subsection{Dataset Generation}
\label{sec:DG}
To demonstrate the effectiveness and scalability of our proposed graph learning framework,
we conduct comprehensive experiments on real-world analog circuits.
Specifically, {40} circuits are designed by experienced analog engineers under commercial 180$nm$ technology, covering various functional types,
{including operational transconductance amplifier (OTA), comparator (COMP), bandgap reference circuit, low dropout regulator circuit (LDO) and oscillator circuit, which are widely used in analog front-end applications.}
The OTA circuit is commonly employed as a signal-processing component. %is widely used in integrated amplifier and filter applications.
%serving a range of applications including amplification, filtering, and integration.
%Their amplification factor can be flexibly adjusted to meet specific design requirements.
The COMP circuit facilitates comparison between analog and reference signals,
subsequently generating digital signals or switching control signals based on the comparison results.
%Due to the characteristics of high speed and common mode rejection ratio (CMRR),
The COMP circuit is valued for its high speed, accuracy, and common mode rejection ratio (CMRR),
and its device parameters significantly impact the performance of the entire circuit.
A bandgap reference circuit is designed to provide PVT (process, supply voltage, temperature) insensitive reference voltages,
which is a critical component in analog circuit design.
{The LDO circuit is a linear regulator that controls high voltage to a stable lower voltage.
The low dropout characteristic makes LDO circuits useful in applications that require high output voltage accuracy, a wide range of input voltage, and low power consumption,
such as mobile devices, wireless communication, and embedded systems.}
{The oscillator circuit is used to generate periodic electrical signals, such as sine waves, square waves, pulses, etc.
These electrical signals are commonly employed as clock signals, frequency references, modulation/demodulation, communication systems, and timing control in various electronic devices.}
%\textcolor{blue}{These are all common circuits in circuit design. }

Based on these {40} circuits,
two datasets are constructed, the first of which contains all circuits named Hybrid,
while the second includes only the OTA circuits denoted OTA.
As described in \Cref{sec:dgr}, the SPICE circuit netlist can be represented as a directed graph,
where devices and IO ports are nodes, and the nets are recognized as edges.
%We extract all the necessary information from the SPICE files to generate graph feature embeddings.
%The valid pairs are potential constraint candidates as defined in~\cite{gao2021layout}.
%In our dataset, we define a valid pair as one that matches the type rule.
In order to avoid a match between the nodes of the different kinds of circuits,
a separate graph for each circuit is constructed to form the training set.
Note that we only label symmetrical devices within each circuit graph as valid device pairs.
In our dataset, we define a valid pair as one that matches the type rule.
Besides, for devices with extra pins (e.g., triple-well devices), we also require that the corresponding valid pair should have the same potential.
We roughly divide each dataset into training and testing sets in a 3:1 ratio at the circuit level.
%When each circuit is tested, the machine learning model is trained only on the training sets.
The graph learning-based model is trained only on the training set.
This ensures that the tested circuits are totally unseen to the network model,
eliminating the possibility of information leaking from the testing set to the training set.
Notably, the training time of the proposed EGAT model on the Hybrid and OTA datasets is only {128.211$s$} and 57.596$s$ respectively,
indicating a very rapid training process.
In addition, the testing set of the Hybrid dataset encompasses all types of circuits.
%At the same time, it also ensures the good generalization of the model on other unseen circuits.
The statistics on the circuit graphs are listed in \Cref{table:2}. %lists the a recapitulation of our collected datasets.
%Then we extract all the necessary information from the SPICE files to generate graph feature embeddings.

%These circuits are widely used in practical analog circuits, and are a critical component in any analog circuit design, motivating their inclusion within our datasets.
%Based on the analog circuit,
%the corresponding SPICE circuit netlist is represented as a directed graph as shown in \Cref{sec:dgr},
%where devices and IO ports are vertices, and the nets are recognized as edges.
%We extracted all necessary information from the SPICE files pertaining to these circuits to generate graph feature embeddings.

%into nodes, while the connections linking these devices and IOs were transformed into edges within the graph.
%Effective pairs of nodes served as candidates for symmetric constraints in the graph, deriving from the combination of device nodes.
%Out of these effective pairs, true symmetric constraint pairs were generated.
%\Cref{table:2} provides a recapitulation of our collected datasets. We integrated all 20 circuits into a single, large circuit to evaluate the model's performance on hybrid circuits, naming it ALL.
%Similarly, we integrated all OTA circuits into one large circuit to evaluate the model's performance in a single-circuit scenario.
%For each large circuit, we divided the datasets into training and testing sets in a ratio of roughly 3:1, ensuring that no information leaked from the testing set to the training set.

\begin{table}
\centering
\renewcommand{\arraystretch}{1.2}
\caption{Statistics of the datasets.}
\resizebox{8.5cm}{!}{
    \begin{tabular}{c|c|c|c|c|c}
    \toprule
    Dataset & \#Graphs  & \#Nodes & \#Edges & \#Valid pairs & \#Matched pairs\\
    \midrule
    Hybrid   & {40}  & {1378} & {10824} & {6020} & {821}\\
    \midrule
    OTA      & 15  & 552 & 3512 & 3153 &358\\
    \bottomrule
    \end{tabular}
    }
    \label{table:2}
\end{table}

\begin{table*}
\centering
\caption{Comparison with state of the arts on Hybrid and OTA datasets (time unit: $s$).}
\renewcommand{\arraystretch}{1.1}
\resizebox{15cm}{!}{
\begin{tabular}{c|c|c|c|c|c|c|c}
\toprule
Dataset & Model &TPR &FPR &PPV &ACC &F$_1$-score  &Inference time\\
\midrule
\multirow{5}{*}{Hybrid}
& SFA~\cite{xu2019magical}& {0.4023} & {0.0081}  & {0.8607} & {0.9267}  & {0.5483} & {\textbf{0.0389}}\\
&ASPDAC'21~\cite{gao2021layout} & {0.2958} & {\textbf{0.0022}}  & {0.9433} & {0.9202}  & {0.4504} & {0.0799}\\
& DAC'21~\cite{chen2021universal} &{0.4437} & {0.0139}  & {0.7979} & {0.9260}  & {0.5703} & {0.0574}\\
& EGAT(w/o. PP)& {\textbf{0.9463}} & {0.0628}  & {0.7289} & {0.9384}  & {0.8048} & {0.0477}\\
& EGAT & {\textbf{0.9463}} & {0.0054}  & {\textbf{0.9481}} & {\textbf{0.9908}}  & {\textbf{0.9444}} & {0.0559}\\
\midrule
\multirow{5}{*}{OTA}
& SFA~\cite{xu2019magical}& {0.3750} & 0.0091  & 0.8295 & 0.9257  & 0.5163 & \textbf{0.0296}\\
&ASPDAC'21~\cite{gao2021layout} & 0.2682 & \textbf{0.0013} &\textbf{0.9642} & 0.9135  & 0.4196  &0.1136\\
&DAC'21~\cite{chen2021universal} & 0.3600 & 0.0218  & 0.7500 & 0.8833  & 0.4865  &0.0993\\
& EGAT(w/o. PP)& \textbf{1.0000} & 0.0206  & 0.8502 & 0.9815  & 0.9166 & 0.0758\\
& EGAT & \textbf{1.0000} & 0.0047  & 0.9629 & \textbf{0.9958}  & \textbf{0.9807}  &0.0941\\
\bottomrule
\end{tabular}
}
\label{table:4}
\end{table*}

\begin{table}
\centering
\renewcommand{\arraystretch}{1.2}
\caption{{The FPR and TPR values of EGAT(w/o. PP) at different thresholds (Hybrid dataset).}}
\resizebox{8.5cm}{!}{
    \begin{tabular}{c|c|c|c|c|c|c}
    \toprule
    Threshold   & 0.5 & 0.6 & 0.7 & 0.8 & 0.9 &0.99\\
    \midrule
    FPR   & 0.0908 & 0.0628 & 0.0450 &0.0365 &0.0214 &0.0070\\
    \midrule
    TPR   & 0.9565 & 0.9463 & 0.9318 &0.9018 &0.8571 &0.3014\\
    \bottomrule
    \end{tabular}
    }
    \label{table:threshold}
\end{table}

\subsection{Baselines Description}

To verify the superiority of the proposed framework,
we compare our framework with three state-of-the-art symmetric constraint detection methods, including
{SFA~\cite{xu2019magical}, ASPDAC'21~\cite{gao2021layout} and DAC'21~\cite{chen2021universal}.
%We compare our model with conducted a comparative analysis of our model with two other extensively used symmetric constraint detection methods for analog circuits, namely $S^{3}DET$~\cite{liu2020s} and Layout~\cite{gao2021layout}.
%Originally, S$^3$DET focuses on the extraction of the system-level symmetry constraints.
%Thanks to the modifications by Lin \textit{et al}.~\cite{gao2021layout},
%the algorithm adapts to the device-level symmetry constraint detection,
%subsequently renaming it S$^3$DET-dl.
%The S$^3$DET~\cite{liu2020s} algorithm leverages spectral graph analysis and graph centrality to detect system-level symmetry constraints,
%but it is equally applicable to device-level symmetry constraint extraction,
%so we subsequently renaming it S$^3$DET-dl.
The SFA~\cite{xu2019magical} abstracts the SPICE file into a graph representation. Following this, seed pattern detection is employed to examine the connection relationships and device attributes between the pairs of devices for matching. This process yields seed symmetric pairs, which serve as a starting point for traversing the graph to identify the remaining symmetric pairs.
In ASPDAC'21~\cite{gao2021layout}, a GraphSAGE model is exploited to aggregate adjacent information into node embeddings,
which are adopted to identify whether two nodes are symmetric.
Besides, the DAC'21~\cite{chen2021universal} presents a graph learning-based framework leveraging unsupervised learning to recognize circuit matching structures.

\subsection{Comparison with Baselines}
\label{sec:cwb}

We compare our framework with \textcolor{blue}{SFA~\cite{xu2019magical}}, ASPDAC'21~\cite{gao2021layout} and DAC'21~\cite{chen2021universal} on the two datasets.
In the experiment, we distinguish the EGAT model without and with post-processing rules as ``EGAT(w/o. PP)'' and ``EGAT'', respectively.
To ensure a fair comparison, all the parameters of the SOTA works are set the same as those in these original papers.
%Note that the parameters values used in S$^3$DET-dl and ASPDAC'21 are the same as in the published papers.
\Cref{table:4} lists the experimental results.
The evaluation metrics TPR, FPR, PPV, ACC, and F$_1$-score are defined in \Cref{sec:pf}.
%Columns ``Training time'' and ``Inference time'' represent the %total training time and the average inference time per circuit.
%Column ``Training time'' represents the total training time on all training circuits.
Meanwhile, column ``Inference time'' denotes the average inference time on each tested circuit, which also contains the post-processing time.
We emphasize the better results in bold.
As shown in the table, the proposed framework significantly outperforms {SFA~\cite{xu2019magical}}, ASPDAC'21~\cite{gao2021layout} and DAC'21~\cite{chen2021universal} in almost all metrics. %by a large margin.
%We summarized and compared the experimental results of our proposed framework with those of two benchmark models in \Cref{table:4},
%reporting consistently superior performance across all metrics, including TPR, FPR, PPV, ACC, and $F_{1-score}$.
%When comparing the best values for each indicator with those of $S^{3}$DET-dl and layout, all indicators exhibited higher performance when using our proposed framework in both the ALL and OTA circuits.
For example,
compared with DAC'21~\cite{chen2021universal},
%the proposed EGAT improves the TPR, FPR, PPV, ACC, and F$_1$-score by $19.49\%$, $12.89\%$, $49.77\%$, $13.68\%$, and $38.25\%$, respectively, on the Hybrid dataset.
the proposed EGAT improves the TPR, PPV, ACC, and F$_1$-score by {$50.26\%$, $15.02\%$, $6.48\%$, and $37.41\%$, respectively,
while the FPR is reduced by $0.85\%$ on the Hybrid dataset.}
%we achieved an increase in TPR by $28.58\%$, while decreasing FPR by $99.49\%$, increasing PPV by $122.08\%$, and boosting ACC and $F_{1-score}$ by $17.68\%$.
Meanwhile,
on the OTA dataset,
the proposed EGAT framework achieves $99.58\%$ ACC and $98.07\%$ F$_1$-score.
%Our model's training times on Hybrid and OTA are 84.762s and 57.596s respectively, indicating a very rapid training process.
The reason is that in SFA~\cite{xu2019magical} and ASPDAC'21~\cite{gao2021layout},
a limitation is imposed wherein each device can have at most one symmetry constraint.
For instance, if devices $v_1$ and $v_2$ are a symmetric pair, $v_1$ and $v_3$ will never be symmetric.
This limitation leads to a degradation of performance metrics, which is also why these works get a lower FPR.
Besides, DAC'21~\cite{chen2021universal} does not account for devices other than MOSFETs and passive devices.
Hence, the performance of DAC'21 is poor in our datasets, which also contain NPN and PNP devices.
In contrast, our proposed EGAT model effectively addresses these limitations, resulting in improved performance metrics. 
%Moreover, As for the decline in performance of EGAT(w/o. PP) compared to EGAT is primarily attributed to two factors. Firstly, the post-processing step significantly enhances the overall results. Secondly, as mentioned earlier, setting the threshold to 0.6 is a deliberate choice considering the post-processing selection, not exclusive to EGAT only. This decision has introduced a certain influence on the performance, too.}
Note that the threshold used in \Cref{7} to calculate the similarity score significantly impacts the final results.
\Cref{table:threshold} provides the TPR and FPR values of EGAT(w/o. PP) under different thresholds.
It can be observed that although higher thresholds result in lower FPR, the TPR is further decreased.
After considering the overall results of various metrics, we adopted a threshold of 0.6 in the experiment.
In addition, despite the higher FPR of EGAT(w/o. PP) in \Cref{table:4}, our post-processing rules can eventually reduce the final FPR.}
%The choice of different thresholds will impact the final results. \Cref{table:threshold} provides the FPR values of EGAT(w/o. PP) at different thresholds. Although higher thresholds result in lower FPR values, there are corresponding losses in other metrics. After considering the overall results of various metrics, we chose 0.6 as the final threshold. Despite the higher FPR of EGAT(w/o. PP) in \Cref{table:4}, our post-processing effectively reduces the FPR, addressing this issue.}

We also evaluate the inference time of the proposed methodology.
Except for SFA~\cite{xu2019magical}, which is compiled in \texttt{C++} and exhibits faster execution time,
the inference time of our EGAT model is less than those of ASPDAC'21~\cite{gao2021layout} and DAC'21~\cite{chen2021universal}.
That is because our graph construction is simpler compared to ASPDAC'21~\cite{gao2021layout} and DAC'21~\cite{chen2021universal}, especially for devices with a large number of pin connections.
Consequently, the reduced parameters decrease the inference time.
%That is because ASPDAC'21~\cite{gao2021layout} incorporates device pins as nodes in the model, which leads to increased parameters,
%consequently prolonging the inference time.
%Moreover, the inference time for DAC'21~\cite{chen2021universal} and our EGAT model are very close, with almost negligible differences.
%The reason for the difference is that our graph construction is simpler compared to DAC'21~\cite{chen2021universal}, especially for devices with a large number of pin connections.}

\begin{figure}[tb!]
	\centering
	% ==== defined colors
%\definecolor{myorange}{RGB}{244,106,18} %F47012
%\definecolor{myblue}{RGB}{0,111,190}    %006FBE
%\definecolor{mygreen}{RGB}{134,182,54}
%\definecolor{myred}{RGB}{228,46,36}     %E42E24
%\definecolor{myyellow}{RGB}{198,148,34} %C69422
%\definecolor{mydark}{RGB}{114,44,114}   %722C72
%\definecolor{mymiddle}{RGB}{144,44,144} %902C90
%\definecolor{mylight}{RGB}{167,44,167}  %A72CA7
\definecolor{myorange}{RGB}{244,106,18} %F47012
%\definecolor{myblue}{RGB}{0,111,190}    %006FBE
\definecolor{myblue}{RGB}{71,162,241}      %47A2F1
%\definecolor{mygreen}{RGB}{134,182,54}
\definecolor{mygreen}{RGB}{142,202,206}    %8ECACE
\definecolor{myred}{RGB}{228,46,36}     %E42E24
%\definecolor{myyellow}{RGB}{198,148,34} %C69422
\definecolor{myyellow}{RGB}{255,253,181}   %FFFDB5
\definecolor{mydark}{RGB}{114,44,114}   %722C72
\definecolor{mymiddle}{RGB}{144,44,144} %902C90
%\definecolor{mylight}{RGB}{167,44,167}  %A72CA7
\definecolor{mylight}{RGB}{191,191,255}
\definecolor{mypink}{RGB}{255,171,164}     %FFABA4
\definecolor{lightblue}{RGB}{221,239,250}  %DDEFFA

\pgfplotsset{
    width =0.40\textwidth,
    height=0.32\textwidth,
    /pgfplots/area legend/.style={
        legend image code/.code={
        \draw[#1] (0cm,-0.1cm) rectangle (0.4cm,0.1cm);
        }
    }
}

\begin{tikzpicture}
	\begin{axis}[
		xmin=0,xmax=1,ymin=0,ymax=1,
        scaled ticks=false,
		xticklabel style={/pgf/number format/fixed,/pgf/number format/precision=1},
		xtick={0,0.2,0.4,0.6,0.8,1.0},
		yticklabel style={/pgf/number format/fixed,/pgf/number format/precision=1},
        ytick={0,0.2,0.4,0.6,0.8,1.0},
		xlabel=FPR,
        ylabel=TPR,
		%xlabel near ticks,
		legend style={
			nodes={scale=0.64, transform shape},
			draw=none,
			at={(0.651,0.32)},
			anchor=north,
			legend columns=1,
		}
		]

		\addplot [color=gray, line width=1pt,mark = *]  table [x={x},  y={y}]  {s3det.dat};
		\addplot [color=mygreen, line width=1pt]    table [x={x},  y={y}]  {aspdac.dat};
		\addplot [color=myblue, line width=1pt]   table [x={x},  y={y}]  {dac.dat};
        \addplot [color=myred, line width=1pt]    table [x={x},  y={y}]  {egat.dat};
        \addplot [/tikz/dashed, color=black, line width=1pt]    table [x={x},  y={y}]  {base.dat};
		\legend{SFA~\cite{xu2019magical}~, ASPDAC'21~\cite{gao2021layout}~(AUC=0.746), DAC'21~\cite{chen2021universal}~(AUC=0.882), EGAT(w/o. PP)~(AUC=0.956)}
		
	\end{axis}
\end{tikzpicture}
	\caption{{The ROC curves of different models (Hybrid dataset).}}
	\label{fig:ROC}
\end{figure}
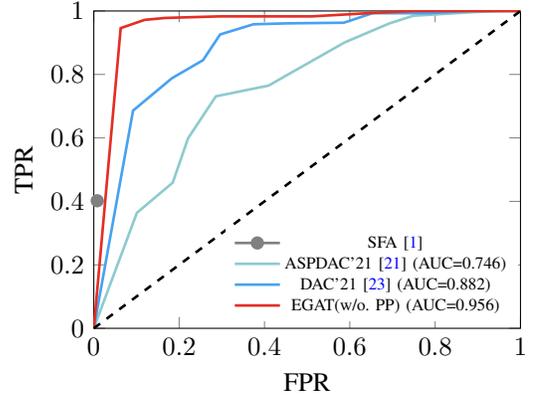

To further demonstrate the performance of the proposed methodology,
%after eliminating all post-processing rules,
we plot the Receiver Operating Characteristic (ROC) curves of our {EGAT(w/o. PP)} model and the three baselines on the Hybrid dataset, respectively.
For the binary classification problem, the ROC curve~\cite{John1973The} is a graphical evaluation tool showing the FPR on the $x$-axis and the TPR on the $y$-axis.
By calculating the Area Under the ROC Curve (AUC), the model's performance at different classification thresholds can be estimated.
%showcasing the model's performance at different classification thresholds by calculating the Area Under the ROC Curve (AUC).
%better showcase our model's performance, after removing all post-processing rules, we plotted the Receiver Operating Characteristic (ROC) curves and compared them with two other baseline models on Hybrid dataset. The ROC curve is a graphical tool used to assess the performance of binary classification models. It depicts the true positive rate on the y-axis and the false positive rate on the x-axis, showcasing the model's performance at different classification thresholds by calculating the Area Under the ROC Curve (AUC).
As depicted in \Cref{fig:ROC},
the solid gray circle represents SFA~\cite{xu2019magical}.
Due to the threshold-independent nature of SFA, it does not allow for adjusting thresholds.
Thus, only one point on the ROC curve for SFA.
The green and blue curves correspond to the ROC curves of ASPDAC'21~\cite{gao2021layout}, and DAC'21~\cite{chen2021universal},
with associated AUC values of {0.746 and 0.882}, respectively.
Meanwhile, the red curve represents the ROC curve of our EGAT(w/o. PP) model,
achieving a high AUC value of {0.956}.
Besides, the black dashed line denotes the baseline with an AUC value of 0.5, similar to the randomized guessing.
In practice, models achieving an AUC greater than 0.8 are commonly regarded as having commendable performance,
while those exceeding 0.9 are deemed excellent classifiers.
Therefore, our EGAT model demonstrates outstanding classification capabilities.

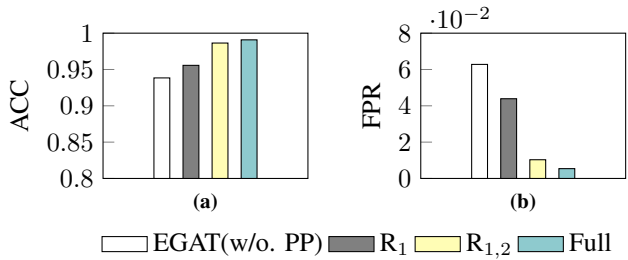
\begin{figure}[tb!]
    \centering
    \begin{tikzpicture}
% \pgfplotsset{
% 	every axis plot/.append style = {font = \tiny}
% }

% ==== defined colors
%\definecolor{myorange}{RGB}{244,106,18} %F47012
%\definecolor{myblue}{RGB}{0,111,190}    %006FBE
%\definecolor{mygreen}{RGB}{134,182,54}
%\definecolor{myred}{RGB}{228,46,36}     %E42E24
%\definecolor{myyellow}{RGB}{198,148,34} %C69422
%\definecolor{mydark}{RGB}{114,44,114}   %722C72
%\definecolor{mymiddle}{RGB}{144,44,144} %902C90
%\definecolor{mylight}{RGB}{167,44,167}  %A72CA7
\definecolor{myorange}{RGB}{244,106,18} %F47012
%\definecolor{myblue}{RGB}{0,111,190}    %006FBE
\definecolor{myblue}{RGB}{71,162,241}      %47A2F1
%\definecolor{mygreen}{RGB}{134,182,54}
\definecolor{mygreen}{RGB}{142,202,206}    %8ECACE
\definecolor{myred}{RGB}{228,46,36}     %E42E24
%\definecolor{myyellow}{RGB}{198,148,34} %C69422
\definecolor{myyellow}{RGB}{255,253,181}   %FFFDB5
\definecolor{mydark}{RGB}{114,44,114}   %722C72
\definecolor{mymiddle}{RGB}{144,44,144} %902C90
%\definecolor{mylight}{RGB}{167,44,167}  %A72CA7
\definecolor{mylight}{RGB}{191,191,255}
\definecolor{mypink}{RGB}{255,171,164}     %FFABA4
\definecolor{lightblue}{RGB}{221,239,250}  %DDEFFA

\begin{groupplot}[
    group style={
        group name=my plots,
        group size=2 by 1,
        horizontal sep=1.5cm
    },
    width=.49\linewidth,
    height=.40\linewidth,
    ybar=5pt,
    enlargelimits=0.05,
    enlarge y limits=0,
    legend style={
        at={(-0.3,-0.3)},
        draw=none,
        anchor=north,
        legend columns=0.2
    },
    area legend,
    symbolic x coords={\texttt{Average}},
    xtick=\empty,
    ytick pos=left,
]
\nextgroupplot[
    bar width=6pt,
    ylabel={ACC},
    ymin= 0.80,
    ymax=1.00,
    yticklabel style={/pgf/number format/fixed},
    ytick={0.80,0.85,...,1.05},
    title style={at={(0.5,-0.4)}},
    title={\footnotesize\textbf{(a)}},
]
    \addplot[fill=white]    coordinates {(\texttt{Average},0.9384)}; 
    \addplot[fill=gray]    coordinates {(\texttt{Average},0.9557)};
    \addplot[fill=myyellow]     coordinates {(\texttt{Average},0.9864)};
    \addplot[fill=mygreen] coordinates {(\texttt{Average},0.9908)};
\nextgroupplot[
    bar width=6pt,
    ylabel={FPR},
    ymin=0.00,
    ymax=0.08,
    title style={at={(0.5,-0.4)}},
    yticklabel style={/pgf/number format/fixed,/pgf/number format/precision=2},
    ytick={0.00,0.02,...,0.10},
    title={\footnotesize\textbf{(b)}},
]
    \addplot[fill=white]    coordinates {(\texttt{Average},0.0628)}; 
    \addplot[fill=gray]    coordinates {(\texttt{Average},0.0439)};
    \addplot[fill=myyellow]     coordinates {(\texttt{Average},0.0103)};
    \addplot[fill=mygreen] coordinates {(\texttt{Average},0.0054)};
    \legend{EGAT(w/o. PP)\ ,R$_1$\ ,R$_{1,2}$\ ,Full }
\end{groupplot}
\end{tikzpicture}
    \caption{{Comparisons of (a) ACC and (b) FPR of different post-processing rules (Hybrid dataset).}}
    \label{fig:ablation1}
\end{figure}

In the layout automation of analog circuits,
constraint extraction is the first stage where higher TPR and lower FPR values can alleviate the difficulties and computational costs of the subsequent placement and routing,
significantly reducing the overall design complexity.
More importantly,
since the circuits vary considerably in Hybrid and OTA datasets,
the experimental results also demonstrate that our framework can generalize to different types of analog circuits.
%Specifically, in the context of automation of analog circuit design, constraint extraction represents the initial step, and higher TPR values and lower FPR values enable the minimization of subsequent layout and routing difficulty and computational cost, leading to a significant reduction in overall design complexity.

\subsection{Ablation Study}
\label{sec:as}

\begin{figure}[tb!]
	\centering
    \hspace{-.17in}
	\includegraphics[width=3in]{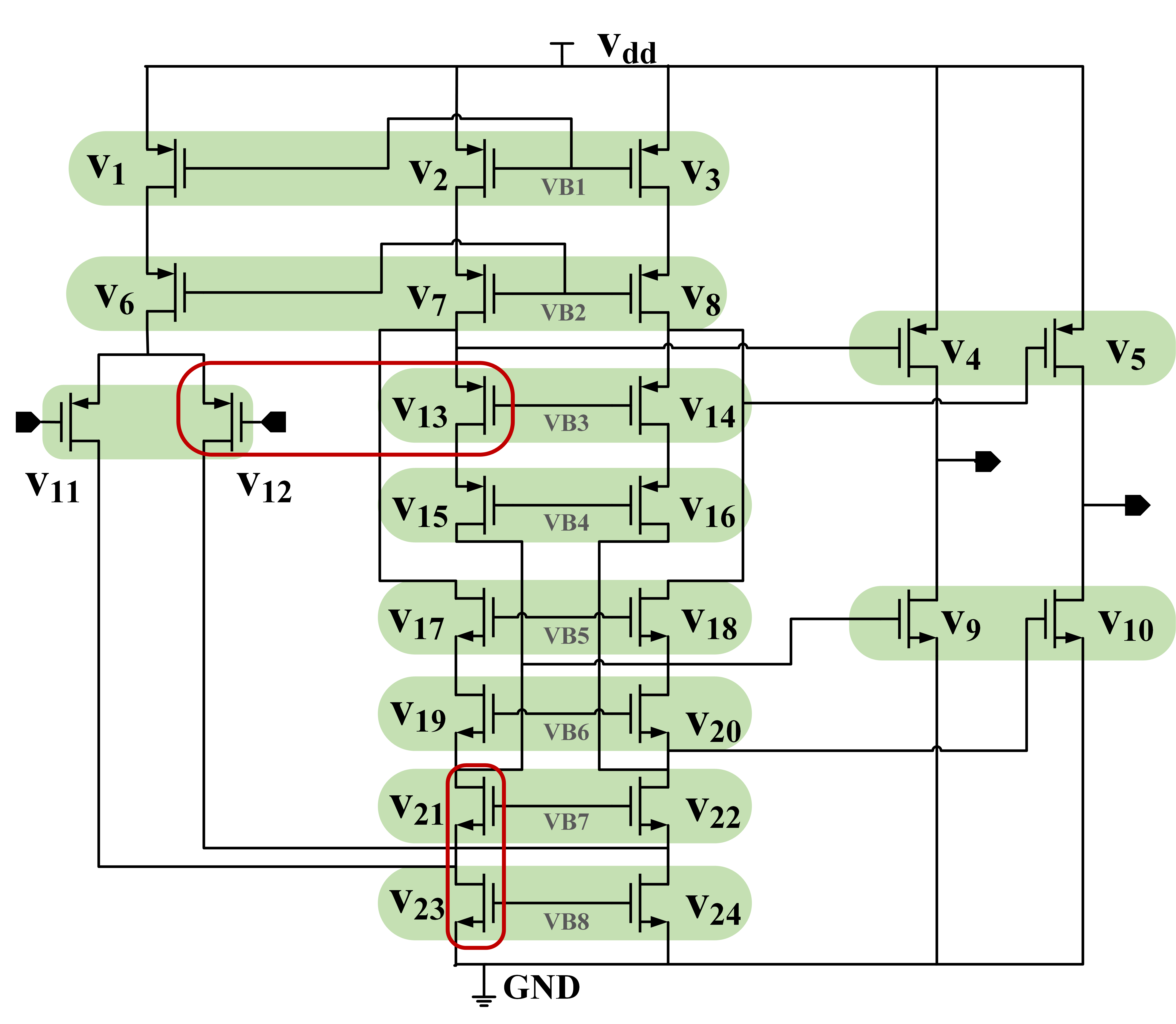}
	\caption{Inference for a Class-AB OTA circuit.}
	\label{fig:classAB}
\end{figure}

An ablation study is performed to investigate how different post-processing rules affect the performance.
\Cref{fig:ablation1} summarizes the contribution of the post-processing rules. 
{``EGAT(w/o. PP)'' represents the framework without any post-processing rules,
``R$_1$'' refers to the framework only with the device position rule,
``R$_{1,2}$'' stands for the framework with the device position and size rules, while ``Full'' is our framework with entire post-processing rules.}
%``R$_{1,2,3}$'' denotes the framework without the dummy-based rule,
{The histogram shows that comparing the framework with only EGAT model, the post-processing rules get 5.24\% further improvement on ACC and reduce 5.74\% of the FPR.}
\Cref{fig:classAB} illustrates the possible inference process for a Class-AB OTA circuit,
with the green color representing symmetric pairs correctly identified by the network,
and the red lines indicating asymmetric pairs incorrectly recognized by the network but removed during post-processing.

\begin{table}[tb!]
\centering
\renewcommand{\arraystretch}{1.2}
\caption{{Comparison with the traditional GAT model on Hybrid dataset.}}
\resizebox{8.5cm}{!}{
    \begin{tabular}{c|c|c|c|c|c}
    \toprule
      & TPR  & FPR & PPV & ACC & F$_1$-score\\
    \midrule
    GAT  & 0.8875    & 0.0926 & 0.5434 & 0.9051   & 0.6741\\
    \midrule
    EGAT(w/o. PP)   & 0.9463  & 0.0628 & 0.7289 & 0.9384 & 0.8048\\
    \midrule
    Improvement   & \textbf{+5.88\%}  & \textbf{-2.98\%} & \textbf{+18.55\%} & \textbf{+3.33\%} & \textbf{+13.07\%} \\
    \bottomrule
    \end{tabular}
    }
    \label{table:gat}
\end{table}

We also compare the proposed EGAT(w/o. PP) model with the traditional GAT model.
In the experiment, both models utilize three graph attentional layers with nearly identical number of parameters.
\Cref{table:gat} lists the experimental results. 
{It can be seen that compared to the traditional GAT model, the proposed EGAT(w/o. PP)  model exhibits significant improvements in TPR, PPV, ACC, and F$_1$-score by 5.88\%, 18.55\%, 3.33\%, and 13.07\%, respectively,
while the FPR is decreased by 2.98\%.}
The reason is that the GAT model only focuses on node-level features and cannot achieve feature interaction between nodes and edges.
As a consequence, suboptimal predictions are produced with more non-matching pairs being incorrectly identified as matching pairs.
%the GAT model, due to its inability to handle edge information, struggles to effectively integrate it with node information. This results in suboptimal predictions, with more non-matching pairs being incorrectly identified as matching pairs, leading to an increase in FP. This explains the poor results in FPR, PPV, and F$_1$-score.}

\begin{figure}[tb!]
    \centering
    \begin{tikzpicture}
% \pgfplotsset{
% 	every axis plot/.append style = {font = \tiny}
% }

% ==== defined colors
%\definecolor{myorange}{RGB}{244,106,18} %F47012
%\definecolor{myblue}{RGB}{0,111,190}    %006FBE
%\definecolor{mygreen}{RGB}{134,182,54}
%\definecolor{myred}{RGB}{228,46,36}     %E42E24
%\definecolor{myyellow}{RGB}{198,148,34} %C69422
%\definecolor{mydark}{RGB}{114,44,114}   %722C72
%\definecolor{mymiddle}{RGB}{144,44,144} %902C90
%\definecolor{mylight}{RGB}{167,44,167}  %A72CA7
\definecolor{myorange}{RGB}{244,106,18} %F47012
%\definecolor{myblue}{RGB}{0,111,190}    %006FBE
\definecolor{myblue}{RGB}{10,89,176}    %006FBE
%\definecolor{mygreen}{RGB}{134,182,54}
\definecolor{mygreen}{RGB}{142,202,206}    %8ECACE
\definecolor{myred}{RGB}{228,46,36}     %E42E24
%\definecolor{myyellow}{RGB}{198,148,34} %C69422
\definecolor{myyellow}{RGB}{255,253,181}   %FFFDB5
\definecolor{mydark}{RGB}{114,44,114}   %722C72
\definecolor{mymiddle}{RGB}{144,44,144} %902C90
%\definecolor{mylight}{RGB}{167,44,167}  %A72CA7
\definecolor{mylight}{RGB}{191,191,255}
\definecolor{mypink}{RGB}{255,171,164}     %FFABA4
\definecolor{lightblue}{RGB}{221,239,250}  %DDEFFA
\definecolor{mygray}{RGB}{102,102,102}
\definecolor{myfav-azure}{RGB}{157,209,207}
\definecolor{myfav-pink}{RGB}{236,169,161}

\begin{groupplot}[
    group style={
        group name=my plots,
        group size=2 by 1,
        horizontal sep=1.7cm
    },
    width=.49\linewidth,
    height=.40\linewidth,
    ybar=5pt,
    enlargelimits=1.3,
    enlarge y limits=0,
    legend style={
        at={(-0.3,-0.3)},
        draw=none,
        anchor=north,
        legend columns=0.2
    },
    area legend,
    %y label style={at={(0.26,0.5)}},
    %symbolic x coords={1,2,3,4,5},
    %xtick=\empty,
    ytick pos=left,
]
%\nextgroupplot[
%    bar width=6pt,
%    ylabel={ACC},
%    %y label style={at={(-0.17,0.5)}},
%    xmin = 1,
%    xmax = 5,
%    xtick={1,2,3,4,5},
%    %xticklabels={1,2,3,4,5},
%    ymin= 0.80,
%    ymax=1.00,
%    yticklabel style={/pgf/number format/fixed},
%    ytick={0.80,0.85,...,1.05},
%    title style={at={(0.5,-0.65)}},
%    title={\footnotesize\textbf{(a)}},
%]
%    \addplot[fill=white]    coordinates {(1,0.8920)};
%    \addplot[fill=gray]    coordinates {(2,0.9613)};
%    \addplot[fill=myyellow]     coordinates {(3,0.9730)};
%    \addplot[fill=lightblue]   coordinates {(4,0.9774)};
%    \addplot[fill=mygreen] coordinates {(5,0.9825)};
\nextgroupplot[
bar width=6pt,
ylabel={ACC},
xmin = 1,
xmax = 5,
xtick={-3,0,3,6.0,9.0},
xticklabels={1,2,3,4,5},
ymin= 0.96,
ymax= 1.00,
yticklabel style={/pgf/number format/fixed},
ytick={0.96,0.97,...,1.00},
title style={at={(0.5,-0.65)}},
title={\footnotesize\textbf{(a)}},
]
\addplot[fill=mylight]    coordinates {(1,0.9763)};
\addplot[fill=myfav-pink]    coordinates {(2,0.9847)};
\addplot[fill=myyellow]     coordinates {(3,0.9908)};
\addplot[fill=lightblue]   coordinates {(4,0.9847)};
\addplot[fill=myfav-azure] coordinates {(5,0.9826)};

%\nextgroupplot[
%    bar width=6pt,
%    ylabel={FPR},
%    %y label style={at={(-0.33,0.5)}},
%    xmin = 1,
%    xmax = 5,
%    ymin=0.00,
%    ymax=0.12,
%    title style={at={(0.5,-0.65)}},
%    yticklabel style={/pgf/number format/fixed,/pgf/number format/precision=2},
%    ytick={0.00,0.03,...,0.12},
%    title={\footnotesize\textbf{(b)}},
%]
%    \addplot[fill=white]    coordinates {(1,0.1162)};
%    \addplot[fill=gray]    coordinates {(2,0.035)};
%    \addplot[fill=myyellow]     coordinates {(3,0.0214)};
%    \addplot[fill=lightblue]   coordinates {(4,0.0154)};
%    \addplot[fill=mygreen] coordinates {(5,0.0094)};
\nextgroupplot[
bar width=6pt,
ylabel={F$_1$-score},
xmin = 1,
xmax = 5,
xtick={-3,0,3,6.0,9.0},
xticklabels={1,2,3,4,5},
ymin=0.80,
ymax=1.00,
title style={at={(0.5,-0.65)}},
yticklabel style={/pgf/number format/fixed,/pgf/number format/precision=2},
ytick={0.80,0.85,...,1.00},
title={\footnotesize\textbf{(b)}},
]
\addplot[fill=mylight]    coordinates {(1,0.9022)};
\addplot[fill=myfav-pink]    coordinates {(2,0.9308)};
\addplot[fill=myyellow]     coordinates {(3,0.9444)};
\addplot[fill=lightblue]   coordinates {(4,0.9313)};
\addplot[fill=myfav-azure] coordinates {(5,0.9216)};
%\addplot[fill=mygreen] coordinates {(1,0.1162)(2,0.035)(3,0.0214)(4,0.9774)(4,0.0154)};
%\addplot[fill=lightblue] coordinates {(1,0.1162)(2,0.035)(3,0.0214)(4,0.9774)(4,0.0154)};
    %\legend{EGAT\ ,R$_1$\ ,R$_{1,2}$\ ,R$_{1,2,3}$\ ,Full }
\end{groupplot}
\end{tikzpicture}
    \caption{{Comparisons of (a) ACC and (b) F$_1$-score of different layer numbers of the EGAT model (Hybrid dataset).}}
    \label{fig:ablation2}
\end{figure}
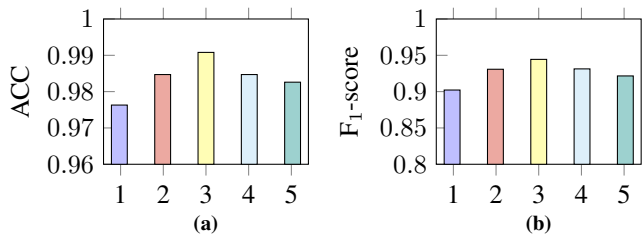

Another ablation study is performed on the graph learning framework to investigate how the number of network layers in the EGAT model affects the performance.
\Cref{fig:ablation2} depicts the changes of ACC and F$_1$-score on the Hybrid dataset with respect to the model's layer numbers.
As illustrated in the histogram, with the network depth increasing, ACC and F$_1$-score get saturated and then decreases gradually.
Thus, it is reasonable to set the layer numbers of the EGAT model to 3.

We further conduct a fourth ablation experiment to demonstrate the impact of the input feature on the final performance.
As described in \Cref{sec:if}, a 15-dimensional node feature is defined,
comprising a 4-dimensional vector that describes the connection relationships among gate ports.
\Cref{table:dim} records the influence of the 4-dimensional gate connectivity vector.
Column ``w/o. GF'' refer to the model without the 4-dimensional gate feature, while ``Full'' is our framework with all defined features. 
{It can be seen that by incorporating the gate port connectivity vector, the TPR, PPV, ACC, and F$_1$-score are enhanced by 2.91\%, 1.54\%, 0.81\%, and 3.58\%,
while the FPR is reduced by 1.42\%,
demonstrating the effectiveness of the defined node features.}
%As shown in \Cref{table:dim}, by incorporating such gate-end connectivity vectors, we successfully increased TPR, PPV, ACC, and F1 by 6.28\%, 3.03\%, 1.03\%, 4.92\%, respectively, and reduced FPR by 0.03\%. This strongly supports the effectiveness of the features we proposed.

\begin{table}[tb!]
\centering
\renewcommand{\arraystretch}{1.2}
\caption{{Comparison of our framework with and without the 4-dimensional gate connectivity feature (Hybrid dataset).}}
\resizebox{8.5cm}{!}{
    \begin{tabular}{c|c|c|c|c|c}
    \toprule
      & TPR  & FPR & PPV & ACC & F$_1$-score\\
    \midrule
    w/o. GF   & 0.9172  & 0.0196 & 0.9327 & 0.9827 &0.9086\\
    \midrule
    Full  & 0.9463  & 0.0054 & 0.9481 & 0.9908 &0.9444\\
    \midrule
    Improvement   & \textbf{+2.91\%}  & \textbf{-1.42\%} & \textbf{+1.54\%} & \textbf{+0.81\%} & \textbf{+3.58\%} \\
    \bottomrule
    \end{tabular}
    }
    \label{table:dim}
\end{table}

	\section{Conclusion}
\label{sec:concl}
In this paper,
we have proposed a graph learning framework to automatically extract symmetric constraints in analog circuits,
%which effectively combines graph edge information with %attentional mechanisms.
which effectively combines edge information in a graph with attentional mechanisms.
Suitable circuit features are designed to guide the network model to achieve information interaction with devices.
Besides, several post-processing rules are developed to significantly reduce the false positive rate.
Experimental results show that, compared with the SOTA implementations, our approach produces better performance.
In the future, we will extend the framework to recognize system-level symmetries in analog circuits. %, which is expected to improve overall levels of automation in analog circuit design.

%We utilized both the device information and circuit connectivity features to develop a comprehensive understanding of symmetric constraint labeling, and incorporated several post-processing methods to further reduce the occurrence of false positives. Our framework demonstrates excellent versatility and broad applicability, making it highly conducive for further promotion and future experimentation. On this basis, we plan to expand the framework's capabilities by exploring its ability to recognize system-level symmetry in circuits, which is expected to improve overall levels of automation in analog circuit design.

    %\clearpage
	
	{
		\bibliographystyle{IEEEtran}
		\bibliography{./ref/Top,./ref/Xuqi,./ref/Software,./ref/DFM,./ref/DL}
	}

\end{document}